\RequirePackage{fix-cm}
\documentclass[twocolumn]{svjour3}
\journalname{International Journal of Computer Vision}

\smartqed
\usepackage{graphicx}
\usepackage[numbers]{natbib}
\usepackage{multirow}
\usepackage{amsmath}
\usepackage{amssymb}
\usepackage{bbm}
\usepackage{bm}
\usepackage{booktabs}
\usepackage{array}

\usepackage{bm}
\usepackage{listings}
\usepackage{subcaption}
\usepackage[table,svgnames, dvipsnames]{xcolor}
\usepackage{mathtools}

\usepackage{mathrsfs}

\definecolor{myred}{rgb}{1,0,0}
\definecolor{mygreen}{rgb}{0,0.6,0}
\definecolor{mygray}{rgb}{0.5,0.5,0.5}
\definecolor{mymauve}{rgb}{0.58,0,0.82}

\usepackage[ruled,vlined]{algorithm2e}

\usepackage{tabularx, booktabs, makecell}
\newcolumntype{Y}{>{\centering\arraybackslash}m{2.2cm}}
\newcolumntype{C}{>{\centering\arraybackslash}X}

\usepackage{xspace}

\usepackage[utf8]{inputenc}
\usepackage[T1]{fontenc}
\usepackage{booktabs}
\usepackage{amsfonts}
\usepackage{nicefrac}
\usepackage{microtype}

\usepackage{graphicx}
\usepackage{amsmath,amssymb}
\usepackage{caption}
\usepackage{float}
\usepackage[most]{tcolorbox}
\usepackage[hidelinks]{hyperref}
\usepackage{wrapfig}
\usepackage[american]{babel}
\usepackage{csquotes}

\usepackage{bm}
\usepackage{paralist}
\usepackage{multirow}

\usepackage[noend]{algpseudocode}

\newcommand{\eg}{{\em e.g.}}
\newcommand{\ie}{{\em i.e.}}

\usepackage{accents}

\definecolor{Gray}{gray}{0.85}
\definecolor{LightCyan}{rgb}{0.88,1,1}

\usepackage{balance}

\makeatletter
\DeclareRobustCommand\onedot{\futurelet\@let@token\@onedot}
\def\@onedot{\ifx\@let@token.\else.\null\fi\xspace}

\def\eg{\emph{e.g}\onedot} 
\def\ie{\emph{i.e}\onedot}

\makeatother

\usepackage{pifont}

\usepackage[misc]{ifsym}
\usepackage{color, colortbl}
\definecolor{citecolor}{HTML}{0071bc}
\definecolor{tabhighlight}{HTML}{e5e5e5}

\makeatletter
\renewcommand\paragraph{
  \@startsection{paragraph}
  {4}
  {\z@}
  {.5em \@plus1ex \@minus.2ex}
  {-.5em}
  {\normalfont\normalsize\bfseries}
}
\makeatother

\newcounter{validationprompt}
\newenvironment{validationprompt}[1]{%
  \refstepcounter{validationprompt}%
  \begin{tcolorbox}[
    enhanced, sharp corners,
    colback=white, colframe=black,
    colbacktitle=black!8, coltitle=black, colupper=black,
    boxrule=0.6pt, titlerule=0.5pt,
    left=5pt, right=5pt, top=4pt, bottom=4pt,
    toptitle=3pt, bottomtitle=3pt,
    before skip=5pt, after skip=5pt,
    fonttitle=\normalfont\fontsize{8.5}{9.5}\selectfont\bfseries\raggedright,
    fontupper=\normalfont\fontsize{8}{9}\selectfont\raggedright,
    title={Prompt \thevalidationprompt: #1}]
  \setlength{\parindent}{0pt}\setlength{\parskip}{1.5pt}%
  \setlength{\emergencystretch}{2em}%
}{\end{tcolorbox}}

\newcommand{\validationsystemprompt}{%
\textbf{System prompt}\\
You are annotating image--text pairs for dataset validation. Use the supplied concept definition to answer the questions below.
\begingroup
\setlength{\leftmargini}{1.8em}
\begin{enumerate}
\setlength{\itemsep}{2pt}\setlength{\parsep}{0pt}
\item Additional content means objects, scene elements, events, or settings other than \textit{[concept]}. Attributes of the concept itself, such as its color or size, do not count.
\item A mismatch means that the image and caption do not cover the same content: an element is visible but not mentioned, or mentioned but not visually represented. For example, an image showing a cup and an orange paired with ``A cup on a table'' has a mismatch because the caption omits the orange.
\end{enumerate}
\endgroup}

\definecolor{ValidationHumanA}{HTML}{EDF3FA}
\definecolor{ValidationHumanB}{HTML}{EDF6EF}
\definecolor{ValidationModelA}{HTML}{FFF5DF}
\definecolor{ValidationModelB}{HTML}{F8EDF1}
\newcommand{\validationresponses}[1]{%
\par\begingroup
\fontsize{7}{8}\selectfont
\setlength{\tabcolsep}{0.5pt}\renewcommand{\arraystretch}{1.1}%
\setlength{\arrayrulewidth}{0.5pt}\arrayrulecolor{black!65}%
\noindent\begin{tabularx}{\linewidth}{|>{\raggedright\arraybackslash}p{0.19\linewidth}|*{5}{C|}}
\hline
Evaluator & U1 & U2 & O1 & O2 & O3 \\
\hline
#1
\hline
\end{tabularx}\par
\endgroup}

\newcommand{\validationquestions}[1]{%
\textbf{Validation questions}\\
\textbf{U1.} Is \textit{[#1]} present in the image?\\
\textbf{U2.} Does the caption express or unambiguously refer to \textit{[#1]}?\\
\textbf{O1.} Does the image contain objects or scene elements other than \textit{[#1]}?\\
\textbf{O2.} Does the caption contain semantic content beyond \textit{[#1]}?\\
\textbf{O3.} Is there a mismatch between the image and its caption?}

\begin{document}
\sloppy

\title{CMCT-G: Cross-Modal Continual Unlearning with Modality and Task-Level Guidance}

\author{Qian Feng \and 
        Wenhu Zhang \and
        Yiming Wu \and
        Jingzhou Chen \and
        Lina Wei \and
        Jiahua Dong \and
        Chao Zhang \and
        Hanbin Zhao \and
        Hui Qian 
}

\institute{Qian Feng \at
            College of Computer Science and Technology, Zhejiang University, China \\
              \email{fqzju@zju.edu.cn}
           \and
            Wenhu Zhang \at
            College of Engineering, The Hong Kong University of Science and Technology, Hong Kong SAR, China \\
              \email{wenhuzhang98@gmail.com}
           \and
           Yiming Wu \at
            School of Computing and Data Science, The University of Hong Kong, China \\
              \email{yimingwu0@gmail.com}
           \and
           Jingzhou Chen \at
            College of Computer Science and Technology, Nanjing University of Science and Technology, China \\
              \email{jzchen@njust.edu.cn}
           \and
            Lina Wei \at
            School of Computer and Computing Science, Hangzhou City University, China \\
            \email{weilin@hzcu.edu.cn}
            \and
           Jiahua Dong \at
            Mohamed bin Zayed University of Artificial Intelligence, Abu Dhabi \\
              \email{dongjiahua1995@gmail.com}
           \and
            Chao Zhang \at
            College of Computer Science and Technology, Zhejiang University, China \\
              \email{zczju@zju.edu.cn}
           \and
           Hanbin Zhao \at
            College of Computer Science and Technology, Zhejiang University, China \\
              \email{zhaohanbin@zju.edu.cn}
           \and
           Hui Qian \at
            College of Computer Science and Technology, Zhejiang University, China \\
              \email{qianhui@zju.edu.cn}
           \and 
           Hanbin Zhao is corresponding author.
           }

\date{}
\renewcommand{\makeheadbox}{}
\titlerunning{CMCT-G: Cross-Modal Continual Unlearning}

\maketitle

\begin{abstract}
Recent advances in cross-modal pre-training have enabled powerful vision-language models but have also created a growing need for concept-level removal to satisfy ethical and regulatory requirements. Prior machine unlearning methods primarily address unimodal or static settings, whereas real-world applications demand the continual removal of sensitive concepts across both modalities. In this paper, we introduce the task of Cross-Modal Continual Unlearning (CMCU), which involves two key shifts: from \textit{object-centric} to \textit{semantic-centric} datasets, and from \textit{static} to \textit{continual} concept removal. CMCU presents two central issues: identifying concept-specific information within loosely aligned image--text pairs and maintaining stable behavior across sequential unlearning tasks. To address these issues, we propose \textbf{CMCT-G}, a unified framework that provides guidance at both the modality and task levels. The CM-G module enhances concept isolation by leveraging purified cues from complementary modalities through cross-modal attention, while the CT-G module introduces prototype-guided dynamic LoRA composition to support modular and interference-aware forgetting. Extensive experiments across multiple benchmarks show that CMCT-G outperforms representative unlearning and continual unlearning baselines under the proposed CMCU protocols, achieving an improved balance between forgetting and preservation. Code is available at \url{https://github.com/RAIAN08/CMCT-G}.
\end{abstract}

\noindent\textit{\textbf{Notice:} This paper includes examples that contain explicit language, references to sexual content, and potentially sensitive or offensive material. These instances are presented solely for research and do not reflect the authors' views.}


\begin{figure*}[t]
  \centering
   \includegraphics[width=1.0\textwidth]{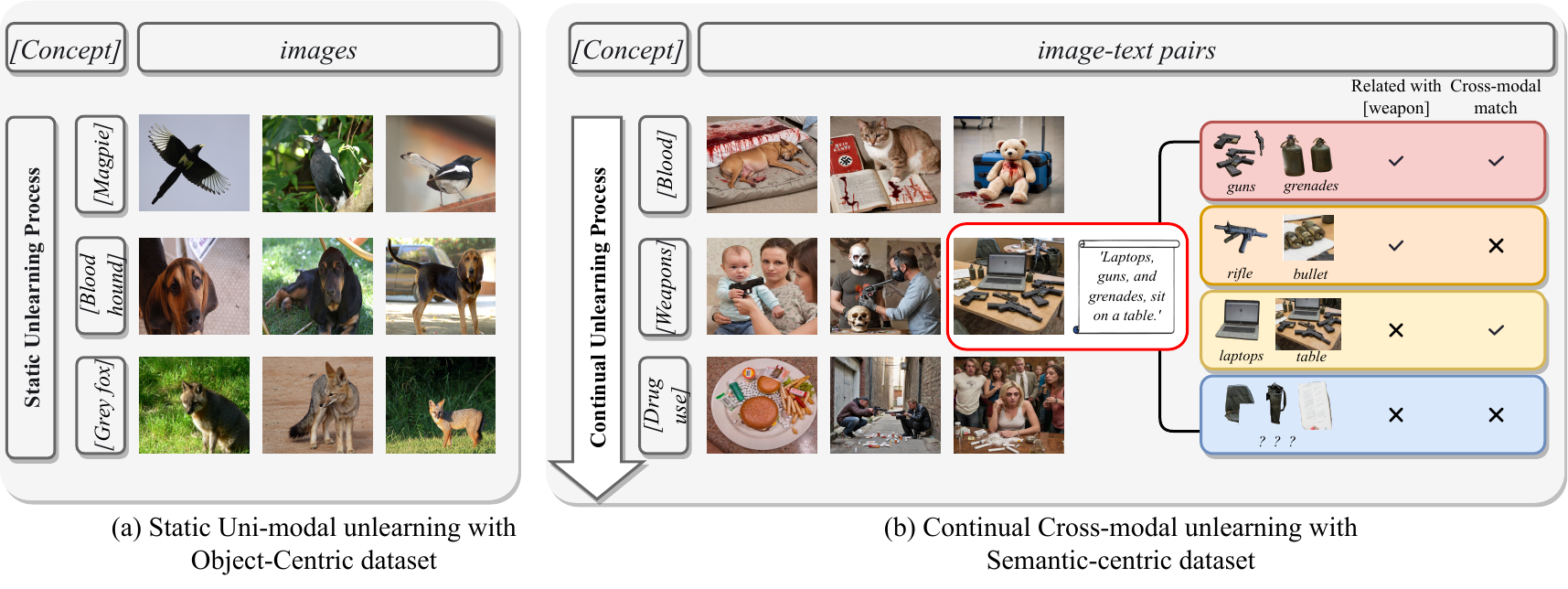}
    \caption{
    Comparison between static and continual unlearning under different \textbf{data structures}. 
    (a) Static unlearning on \textit{object-centric} datasets. Each sample typically contains a single dominant concept with minimal background variation, and the alignment between the visual content and its textual description is direct and unambiguous. Such data make the target concept easy to isolate and minimize interference from unrelated information. 
    (b) Continual unlearning on \textit{semantic-centric} datasets. Each image--text pair naturally includes multiple co-occurring elements, background elements that are not part of the target concept, and occasional mismatches between what appears in the image and what is mentioned in the caption. The target concept may occupy only a small region of the image or a brief portion of the text, and unrelated elements often appear more prominently. These characteristics make selective forgetting substantially more difficult and frequently cause incomplete removal of the target concept or unintended degradation of concepts that should be preserved.
    }
   \label{fig:figure1}
\end{figure*}

\section{Introduction}
\label{sec:intro}

Recent advances in large-scale pre-training have driven the widespread adoption of cross-modal models \cite{radford2021learning, li2022blip, li2023blip, zhai2023sigmoid, zhang2025parameter, han2024parameter, xu2023parameter}, which align vision and text modalities by jointly encoding image--text pairs into a shared representation space. However, the reliance on open and uncurated internet data often introduces undesirable concepts, including biased, harmful, or privacy-sensitive content \cite{nsfwdata, shan2023glaze}. To meet growing demands for ethical, responsible, and trustworthy AI \cite{rosen2011right,regulation2018general}, it is necessary to selectively remove specific concepts (\ie, \textit{forget}) while preserving unrelated concepts (\ie, \textit{retain}). In real-world settings, such removal requests typically arrive over time (\eg, evolving OpenAI safety policies\footnote{https://openai.com/safety/} and EU AI Act\footnote{https://artificialintelligenceact.eu/the-act/}), requiring cross-modal models to continually eliminate concept-specific knowledge while maintaining semantic alignment on unrelated concepts \cite{qureshi2025exploring,zhao2025practical,zhao2024continual,wang2024has,gao2024practical,liu2022continual,chatterjee2024unified,kumaraveluuncle,grimes2024gone}. We refer to this underexplored setting as \textbf{Cross-Modal Continual Unlearning (CMCU)}, a task concerning fairness, privacy, and bias in deployed multimodal systems.

A related line of work is machine unlearning (MU)~\cite{nguyen2022survey, qureshi2025exploring}, which primarily focuses on unimodal architectures such as ResNet and Vision Transformers~\cite{fan2023salun, cheng2024remaining, gao2024practical}. Most approaches operate in a static setting~\cite{zhao2024continual, feng2025fg}, where a fixed concept set is removed once. In contrast, our \textbf{CMCU} setting departs from prior work in two important aspects: the structure of the data (Data Structure) and the nature of the unlearning process (Unlearning Paradigm).

\begin{figure*}[t]
  \centering
  \begin{subfigure}[t]{0.49\textwidth}
    \centering
    \includegraphics[width=\textwidth]{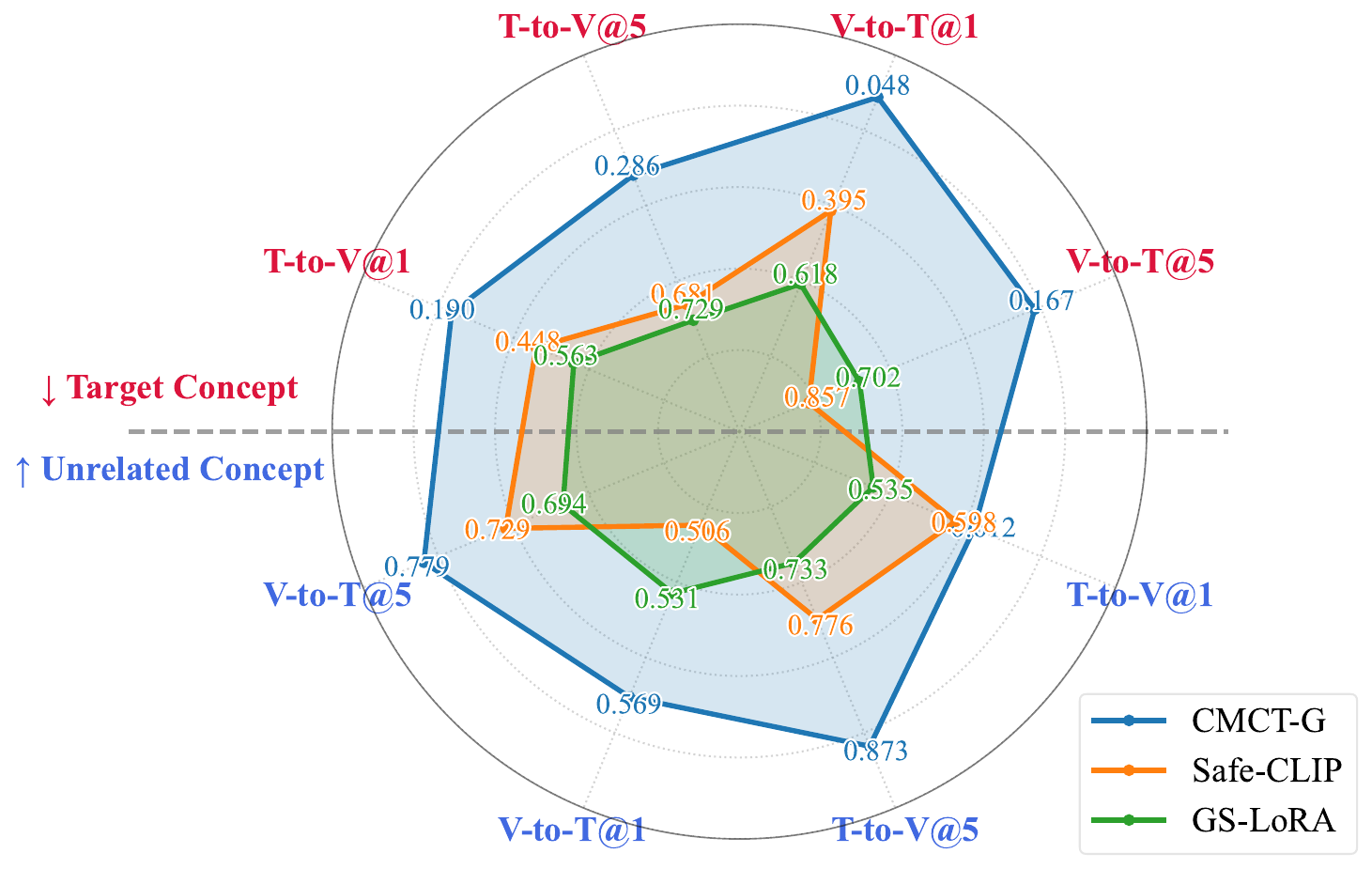}
    \caption{Unlearning a single target concept in a static paradigm under \textit{semantic-centric} datasets.}
    \label{fig:left}
  \end{subfigure}
  \hfill
  \begin{subfigure}[t]{0.49\textwidth}
    \centering
    \includegraphics[width=\textwidth]{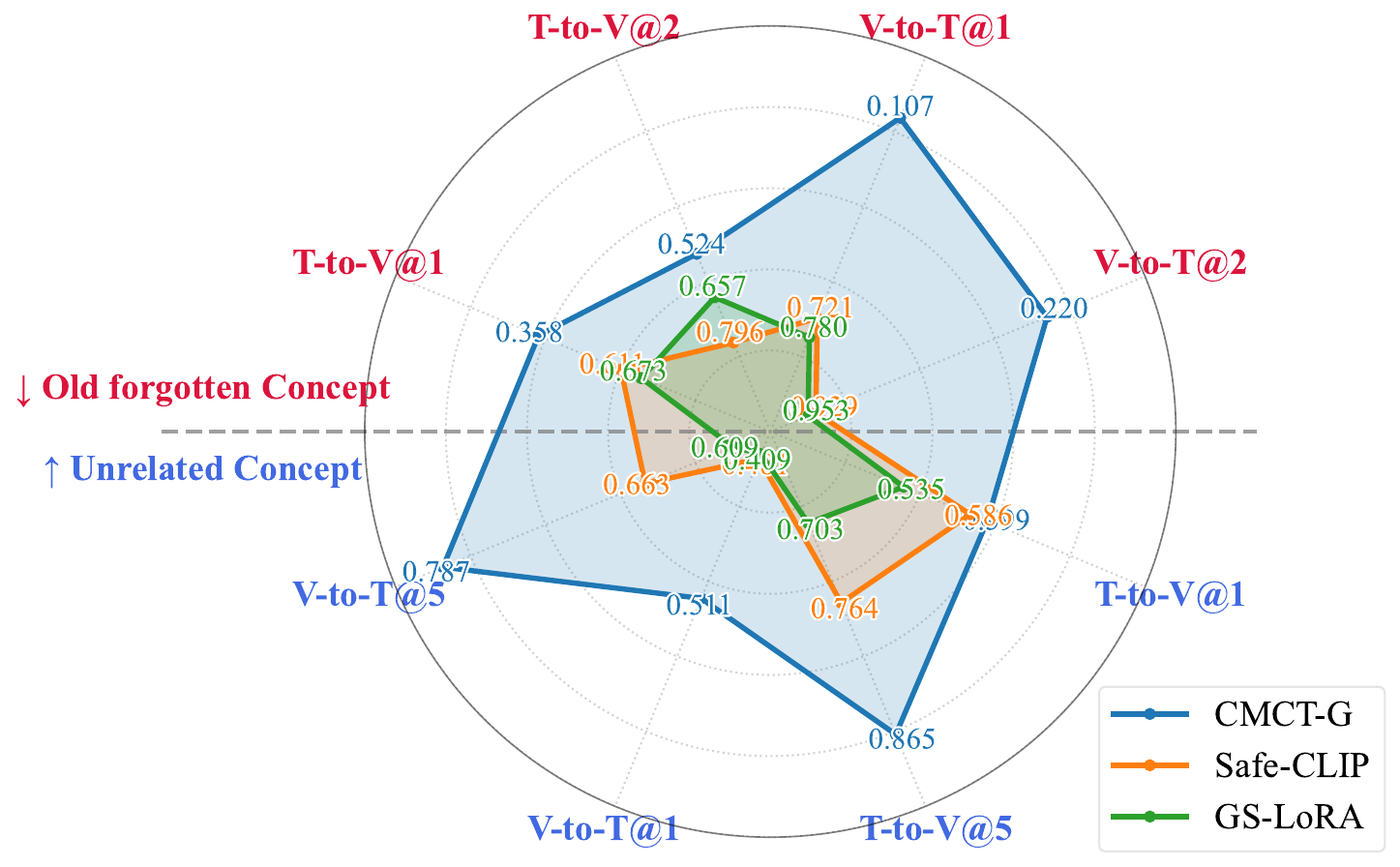}
    \caption{Sequentially unlearning multiple concepts in a continual paradigm under \textit{semantic-centric} datasets.}
    \label{fig:right}
  \end{subfigure}
    \caption{
    Radar charts comparing text-to-vision (\textbf{T}-to-\textbf{V}) and vision-to-text (\textbf{V}-to-\textbf{T}) retrieval across different unlearning methods, evaluated by Recall@$K$. 
    (a) Static unlearning on \textit{semantic-centric} data often leaves residual traces of the target concept and simultaneously suppresses unrelated concepts due to weak image--text alignment. 
    (b) Under continual unlearning, these errors accumulate, previously forgotten concepts reappear, and unrelated semantics deteriorate progressively, revealing the instability of existing MU approaches. 
    Safe-CLIP~\cite{poppi2024safe} and GS-LoRA~\cite{zhao2024continual} are included for reference. 
    \textcolor{black}{$\downarrow$} and \textcolor{black}{$\uparrow$} indicate that lower retrieval scores on forgotten concepts and higher retrieval scores on retained concepts are better, respectively, for both \textbf{T}-to-\textbf{V} and \textbf{V}-to-\textbf{T} retrieval.
    }
  \label{fig:radar}
\end{figure*}

\textbf{Data Structure:} Existing MU methods are typically evaluated on \textit{object-centric} datasets, where each sample contains a single dominant concept and the alignment between image and text is clean and unambiguous. Such data make it easy to delineate what should be forgotten and what should be retained. In our \textbf{CMCU} setting, the data are \textit{semantic-centric} image--text pairs. Samples naturally contain multiple co-occurring elements, backgrounds, and occasional mismatched elements between what appears in the image and what is mentioned in the caption. The target concept often occupies only a small portion of the visual scene or only part of the textual description. These characteristics make it difficult to isolate concept-specific information and frequently lead to incomplete forgetting or unnecessary loss of unrelated concepts, as illustrated in Figure~\ref{fig:radar}(a).

\textbf{Unlearning Paradigm:}
Most MU methods remove concepts in a single pass and do not account for future deletion requests~\cite{caitargeted, poppi2024safe, poppi2025hyperbolic, yang2024cliperase, d2025safe, cheng2024multidelete}. However, in real-world deployments, removal demands arrive over time due to evolving safety policies or legal regulations. Sequential unlearning introduces additional difficulties. Forgetting a new concept may unintentionally revive previously removed concepts, and residual errors may accumulate across tasks. When combined with the ambiguity inherent in \textit{semantic-centric} data, these effects can cause substantial instability, as shown in Figure~\ref{fig:radar}(b).

These observations lead to two central issues that characterize the \textbf{CMCU} problem. (\textbf{Issue 1}) The first issue concerns the ability to locate the exact portion of an image--text pair that encodes the target concept, even when it is surrounded by unrelated content or loosely aligned elements. Errors in this step can lead to over-forgetting or under-forgetting.
 (\textbf{Issue 2}) The second issue arises from sequential removal: the model must incorporate new forgetting tasks without disturbing what has already been removed or degrading concepts that should remain unaffected. Ensuring consistent behavior across tasks is nontrivial, especially when different concepts share overlapping visual or textual cues.

To this end, we propose \textbf{CMCT-G} (\textbf{C}ross-\textbf{M}odality and \textbf{C}ross-\textbf{T}ask \textbf{G}uidance), a unified framework for CMCU that introduces fine-grained guidance at both the {modality level} and the {task level}. At the {modality level}, \textbf{CM-G} (\textbf{C}ross-\textbf{M}odality \textbf{G}uidance) enhances {precise} concept selectivity by injecting signals from the complementary modality via cross-attention. Specifically, in the text-guided image branch in CM-G, we construct \textit{[concept]-only} textual prompts that exclude any unrelated content. These pure text embeddings are used to guide the visual feature extraction process, suppressing irrelevant regions and enhancing concept-specific activations. Symmetrically, in the image-guided text branch in CM-G, we extract localized visual regions for the target \textit{[concept]} using Grounded-SAM \cite{liu2024grounding, kirillov2023segment}. These pure image embeddings are then injected into the textual feature extraction process, enabling the model to downweight semantically unrelated words. These \textit{[concept]-only} prompts and localized visual regions are auxiliary cues derived from the deletion request and are used only during the unlearning stage. By leveraging complementary modality cues, \textbf{CM-G} helps the cross-modal model localize target-related image regions and the corresponding text tokens, thereby addressing \textbf{Issue 1}. At the {task level}, \textbf{CT-G} (\textbf{C}ross-\textbf{T}ask \textbf{G}uidance) mitigates inter-task interference by assigning each deletion task a dedicated LoRA adapter. For each task, we store a prototype summary in the shared representation space and use it to dynamically weight the accumulated LoRA adapters. This routing assigns larger weights to adapters associated with nearby forgetting prototypes and reduces the total LoRA contribution for retained samples far from every stored prototype. By enabling dynamic LoRA weighting and integration of new task modules, \textbf{CT-G} allows CMCT-G to handle sequential deletion requests while reducing inter-task interference, thereby addressing \textbf{Issue 2}.

In summary, our main contributions are as follows:

\begin{itemize}
\item We formulate the \textbf{CMCU} task, which captures the practical need for concept removal in semantic-centric and continually evolving multimodal environments.
\item We construct four corresponding benchmarks for \textbf{CMCU}, including curated datasets, evaluation metrics, and baseline implementations, providing a standardized benchmark suite for this problem.
\item We propose \textbf{CMCT-G}, a unified framework that incorporates cross-modal and cross-task guidance to improve concept localization and to maintain stable behavior across sequential unlearning tasks.
\item Extensive experiments show that \textbf{CMCT-G} achieves a better forgetting-preservation trade-off than representative static unlearning, continual learning, and continual unlearning baselines under the proposed CMCU protocols. Additional analyses on cue robustness, task-order sensitivity, routing alternatives, and efficiency further characterize both its strengths and limitations.
\end{itemize}

\section{Related Work}
\vspace{-1mm}
\subsection{Machine Unlearning for Unimodal Models}

Machine unlearning (MU) aims to remove the influence of specific data from pre-trained models (PTMs) without requiring retraining from scratch~\cite{xu2023machine,zhang2023review}. Existing MU methods are typically categorized into two types: exact unlearning \cite{bourtoule2021machine,liu2021federaser} and approximate unlearning \cite{fan2023salun,kurmanji2023towards}. Despite progress, most existing work focuses on unimodal models such as ResNet and Vision Transformers. These methods are usually evaluated on \textit{object-centric} datasets like CIFAR-100 and ImageNet-1K~\cite{qin2024machine,jia2023model}, which assume that each input image or text exclusively represents the target concept. However, this assumption is overly simplistic and rarely holds in real-world scenarios, where inputs often contain co-occurring content and background information unrelated to the target concept. In contrast, we consider a more realistic setting based on \textit{semantic-centric} datasets, where each image--text pair naturally contains both related and unrelated content. Moreover, image--text pairs often include both matched and mismatched elements, resulting in weak cross-modal alignment. These characteristics introduce new challenges for MU, where imprecise representation of concepts may lead to either under-unlearning or over-unlearning, ultimately compromising model utility and alignment.

\subsection{Machine Unlearning for Cross-Modal Models}

Recent efforts have extended MU to cross-modal models. Some methods~\cite{tu2025sdwv,chin2023prompting4debugging,lu2024mace,gandikota2023erasing} aim to enable MU in generative models, while others~\cite{fan2023salun,yang2024cliperase,poppi2024safe,caitargeted,d2025safe} focus on cross-modal models such as CLIP. However, most existing works assume a \textit{static unlearning} setting, where a single forgetting operation is performed. In contrast, real-world deployments often demand \textit{continual unlearning} to support continuous deletion requests driven by evolving safety policies (e.g., OpenAI’s content guidelines) or regulations such as the EU AI Act. Applying static MU methods in such scenarios often leads to cumulative errors, model collapse, and degraded utility~\cite{wang2024has,gao2024practical,zhao2024continual,feng2025fg}.

Although recent works~\cite{feng2025fg,zhao2024continual} explore MU in continual settings, they still rely on \textit{object-centric} datasets, which assume that each input exclusively represents the target concept. This assumption rarely holds in real-world data, leading to imprecise forgetting and increasing error accumulation, especially under continual settings. These limitations highlight the need for methods that support continual unlearning on \textit{semantic-centric} datasets, aligning with practical cross-modal scenarios. We also note that recent works such as CLIPErase~\cite{yang2024cliperase}, Safe-CLIP~\cite{poppi2024safe}, and Hyperbolic Safe-CLIP~\cite{poppi2025hyperbolic} provide important references for static cross-modal safety unlearning. CLIPErase studies efficient removal of visual-textual associations in CLIP, while Safe-CLIP and Hyperbolic Safe-CLIP focus on safety-aware static unlearning/alignment. Cross-Modal Safety Alignment~\cite{chakraborty2024crossmodal} studies whether textual unlearning can transfer to visual safety, and CLEAR~\cite{dontsov2025clear} studies character-level unlearning in textual and visual modalities. ErrorEraser~\cite{cao2025erroreraser} targets bias removal for continual learning, while recent text-to-image continual unlearning methods~\cite{thakral2025continual,lee2026continual,lee2025empirical} focus on generative diffusion outputs. These works are closely related to ours in motivation, but they differ from our setting in output space, target granularity, optimization objective, or evaluation protocol.

Compared with these settings, CMCU focuses on semantic-centric image--text pairs, sequential concept removal, and bidirectional forgetting-retention evaluation in CLIP-like retrieval models. Later removal requests may interfere with earlier forgotten concepts and retained semantics, motivating both cross-modality localization and task-level stabilization. This setting defines the scope of our CMCU formulation and distinguishes it from static cross-modal unlearning, character-level unlearning, continual learning bias removal, and text-to-image diffusion unlearning.

\begin{figure*}
    \centering
    \includegraphics[width=1\textwidth]{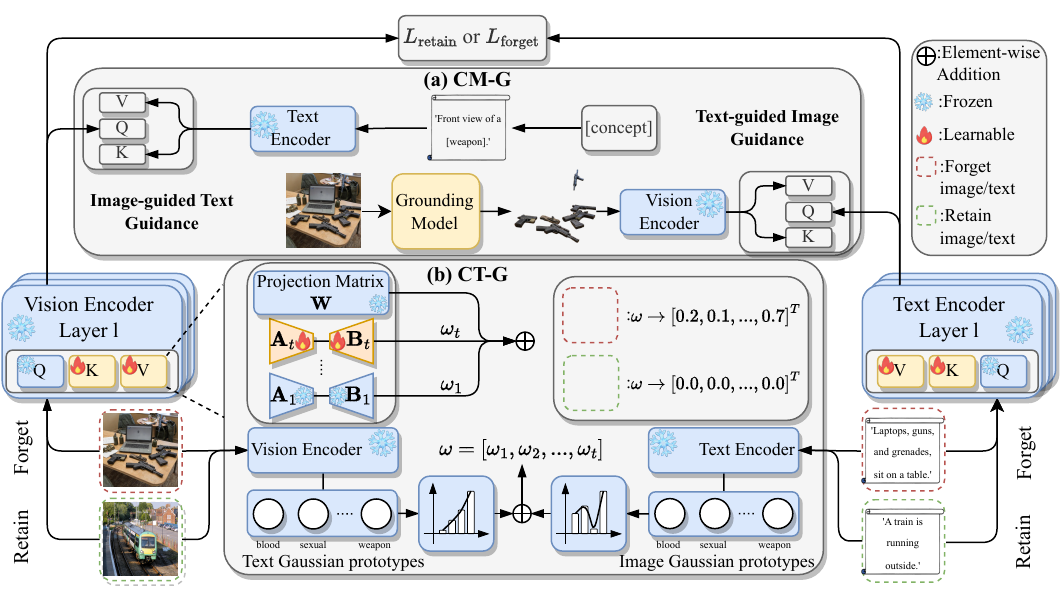}
    \caption{\textbf{Overview of CMCT-G:} (a) For each unlearning concept, we guide the extraction of visual and textual representations by injecting purified signals from the complementary modality. Specifically, \textit{[concept]-only} textual prompts are injected into the vision encoder, while localized visual-region embeddings are injected into the text encoder via cross-attention (Sec.~\ref{Cross-Modality_Guidance}). (b) For a new unlearning task, visual and textual representations are computed as weighted combinations of all LoRA adapter outputs. The weights are obtained by converting cross-modal squared Mahalanobis distances to the stored task prototypes into effective routing weights (Sec.~\ref{Cross-Task_Guidance}).}

    \label{fig:main_fig}
\end{figure*}

\section{Problem Statement}
\label{sec:Preliminary}
\vspace{-3mm}
We formalize the CMCU task, which aims to iteratively remove specified concepts from a cross-modal model (e.g., CLIP) while preserving its alignment performance on unrelated concepts.

Let $\mathcal{M}$ be a pre-trained cross-modal model, and consider a sequence of concepts $\{\mathcal{C}_1, \mathcal{C}_2, \ldots, \mathcal{C}_T\}$ that are intended to be removed over time. For each concept $\mathcal{C}_t$, we construct a {forget set} $\mathcal{D}_{f,t} = \{(v_i, t_i)\}_{i=1}^{N_{f,t}}$ consisting of image–text pairs associated with $\mathcal{C}_t$, where $v_i$ and $t_i$ denote the image and text, respectively, and $N_{f,t}$ represents the number of image–text pairs corresponding to concept $\mathcal{C}_t$. In parallel, we define a {retain set} $\mathcal{D}_{r} = \{(v_j, t_j)\}_{j=1}^{N_{r}}$ composed of image–text pairs from unrelated concepts that should be preserved, where $N_{r}$ denotes the number of such unrelated pairs.

Starting from $\mathcal{M}_0 = \mathcal{M}$, the unlearning algorithm $\mathscr{F}$ iteratively processes each task to produce an updated model $\mathcal{M}_t = \mathscr{F}(\mathcal{M}_{t-1}, \mathcal{D}_{f,t}, \mathcal{D}_{r})$. 

Considering that $\mathcal{M}$ is a cross-modal model, we formulate the forgetting and retaining objectives through a \textbf{cross-modal retrieval} perspective. Specifically, when a target concept $\mathcal{C}_u$ is forgotten, the unlearned model $\mathcal{M}_u$ should be unable to retrieve its paired modality within the corresponding forget set $\mathcal{D}_{f,u}$. That is, given any image $v_i \in \mathcal{D}_{f,u}$, the unlearned model should fail to retrieve its paired text $t_i \in \mathcal{D}_{f,u}$, and vice versa, indicating complete unlearning of the target concept $\mathcal{C}_u$. Meanwhile, for all unrelated concepts, the unlearned model should maintain image–text alignment as before, \ie, given any image $v_i \in \mathcal{D}_{r}$, the model should still retrieve its corresponding text $t_i$, and given any text $t_i \in \mathcal{D}_{r}$, it should retrieve its corresponding image $v_i$.

Formally, the total unlearning process satisfies the following constraints:

\begin{equation}
\left.
\begin{aligned}
\forall v \in \mathcal{F}_{\leq t},\quad
& \arg\max_{t' \in \mathcal{A}_{\leq t}} f_{\mathcal{M}_t}(v, t') \in \mathcal{D}_{r}, \\
\forall t \in \mathcal{F}_{\leq t},\quad
& \arg\max_{v' \in \mathcal{A}_{\leq t}} f_{\mathcal{M}_t}(v', t) \in \mathcal{D}_{r}
\end{aligned}
\right\} \quad \text{(remove)}
\end{equation}

\begin{equation}
\left.
\begin{aligned}
\forall (v, t) \in \mathcal{D}_{r},\quad
& \arg\max_{t' \in \mathcal{D}_{r}} f_{\mathcal{M}_t}(v, t') = t, \\
& \arg\max_{v' \in \mathcal{D}_{r}} f_{\mathcal{M}_t}(v', t) = v
\end{aligned}
\right\} \quad \text{(preserve)}
\end{equation}

where
\[
\mathcal{F}_{\leq t} := \bigcup_{i=1}^{t} \mathcal{D}_{f,i}, \quad
\mathcal{A}_{\leq t} := \mathcal{F}_{\leq t} \cup \mathcal{D}_{r}.
\]

\section{Methods}
\label{sec:methods}

We propose \textbf{CMCT-G}, a unified framework for Cross-Modal Continual Unlearning (CMCU). CMCT-G integrates fine-grained guidance at both the modality and task levels to enable precise removal of specified concepts, particularly under \textit{semantic-centric} datasets where image–text alignment is diverse and weak. An overview of the framework is illustrated in Figure~\ref{fig:main_fig}. {Algorithms~\ref{alg:CMCT-G-train} and~\ref{alg:CMCT-G-infer} describe the continual unlearning and retrieval procedures, respectively.}

\subsection{Cross-Modality Guidance}
\label{Cross-Modality_Guidance}
Accurate localization and representation of target concepts are important for CMCU. Inaccurate localization or representations may result in incomplete or excessive unlearning, leading to residual leakage or unintended removal. Such errors are prone to accumulation under continual settings.

To address this, we introduce \textbf{Cross-Modality Guidance} (CM-G), which consists of two complementary modules: \textbf{text-guided image guidance} and \textbf{image-guided text guidance}. By injecting purified signals from one modality into the representation space of the other, CM-G enables precise cross-modality guidance that strengthens the model’s ability to accurately localize the regions in the raw image most related to the target concept, as well as the specific words in the raw text that correspond to it, thereby providing reliable anchors for downstream unlearning without disrupting unrelated concepts.

\paragraph{Text-guided Image Guidance}
Prior studies \cite{li2025one, poppi2024safe} have shown that visual representations in cross-modal models are often redundant and entangled with both background context and co-occurring content that may be unrelated to the target concept. For instance, images labeled with the concept \textit{[weapon]} may contain unrelated content such as tables or electronic devices, while their paired texts may include elements that are only partially matched or even mismatched with the visual scene. These factors hinder the model’s ability to accurately localize the target concept. To enhance visual selectivity, we construct \textit{[concept]-only} textual prompts (e.g., \textit{[a close-up of a weapon]}) and encode them using a frozen text encoder to obtain purified textual signals from the text modality.

Following \cite{li2025one}, we inject the purified textual signals into the visual feature extraction process via cross-attention, computed as follows:
\begin{equation}
\label{i2v}
\mathbf{\hat{h}}_v^{l} = \text{Softmax}\left(\frac{\mathbf{h}_v^l (\mathbf{e}_t)^\top}{\sqrt{d}}\right) \mathbf{e}_t.
\end{equation}
Here, $\mathbf{h}_v^l$ denotes the raw visual features from the $l$-th layer of the vision encoder, which may contain content unrelated to the target concept or entangled with co-occurring background information. $\mathbf{e}_t$ represents the concept-purified textual embedding used to provide cross-modal guidance. $d$ is the dimensionality of the representation space. Through cross-attention, $\mathbf{e}_t$ acts as a purified signal that strengthens modality interaction and helps align matched elements across modalities while suppressing unrelated visual content and mismatched elements.

The guided visual features $\mathbf{\hat{h}}_v^{l}$ are then fused with the raw visual features through a residual connection:
\begin{equation}
\label{i2v_res}
\mathbf{h}_v^l = \mathbf{h}_v^l + \alpha_v \mathbf{\hat{h}}_v^{l}.
\end{equation}
Here, $\alpha_v$ is the coefficient and is set to $0.5$ by default. The above injection is performed at each layer of the visual feature extraction process, allowing the purified textual signals to progressively deepen their influence on visual representation learning. As a result, the model becomes increasingly capable of emphasizing regions in the image that are semantically related to the target concept, mitigating interference from unrelated content, and reducing the influence of mismatched elements across the image–text pair.

\paragraph{Image-guided Text Guidance}
Textual descriptions are often abstract and may omit concrete visual details that are crucial for accurate concept localization. To address this, we employ a grounded segmentation model, \ie, Grounded-SAM \cite{liu2024grounding, kirillov2023segment}, to extract image regions corresponding to the queried concept (e.g., \textit{[weapon]}). These segmented regions are then encoded using the frozen vision encoder to obtain a concept-purified visual embedding $\mathbf{e}_v$, which serves as purified guidance from the image modality.

We inject the visual guidance embedding into the text encoder via cross-attention at each transformer layer. Given the textual features $\mathbf{h}_t^l$ at layer $l$, the guided representation is computed as:
\begin{equation}
\label{v2i}
\mathbf{\hat{h}}_t^{l} = \text{Softmax}\left(\frac{\mathbf{h}_t^l (\mathbf{e}_v)^\top}{\sqrt{d}}\right) \mathbf{e}_v,
\end{equation}
and integrated with the original representation through a residual connection:
\begin{equation}
\label{v2i_res}
\mathbf{h}_t^l = \mathbf{h}_t^l + \alpha_t \mathbf{\hat{h}}_t^{l}.
\end{equation}
Here, $\alpha_t$ is the coefficient and is set to $0.5$ by default. The concept-purified visual embedding is denoted by $\mathbf{e}_v$, and $d$ is the dimensionality of the representation space. This injection is applied at every layer of the text encoder, enabling the visual guidance to progressively deepen its influence on textual feature learning. Consequently, the textual representation becomes more aligned with visually matched elements, while unrelated textual content or mismatched textual elements are suppressed. This process enhances the model’s capacity to accurately identify and erase the target concept without impairing unrelated semantics.

Grounded-SAM is used only to construct visual guidance cues during the unlearning stage. Once the LoRA adapters and task prototypes have been learned, inference does not require running Grounded-SAM unless a new deletion request arrives. Contextual concepts such as \textit{[harassment]} or \textit{[hate]} may not always correspond to a stable localized region. \textbf{When Grounded-SAM returns no mask for the queried concept, the raw image is used as the visual cue for that training sample.} The noisy-mask sensitivity and qualitative failure cases are analyzed in Sec.~\ref{sec:ablation}.

\begin{algorithm}[!t]
\small
\SetAlgoLined
\caption{{Training procedure of CMCT-G}}
\label{alg:CMCT-G-train}
\KwIn{{Frozen pre-trained CLIP encoders $E_v^0,E_t^0$; deletion concepts $\{\mathcal{C}_t\}_{t=1}^{T}$; forget sets $\{\mathcal{D}_{f,t}\}_{t=1}^{T}$; retain set $\mathcal{D}_r$; coefficients $\alpha_v,\alpha_t,\lambda$.}}
\KwOut{{Task adapters $\{\mathrm{LoRA}_i\}_{i=1}^{T}$ and prototypes $\{\mathcal{P}_i\}_{i=1}^{T}$.}}
\BlankLine
\For{{$t=1$ \KwTo $T$}}{
    {Initialize $\mathrm{LoRA}_t$; keep earlier adapters frozen}\;
    {Extract CM-G features $\{(\mathbf{z}_k^v,\mathbf{z}_k^t)\}_{k=1}^{n}$ from $\mathcal{D}_{f,t}$ using concept $\mathcal{C}_t$ via Eqs.~\ref{i2v}--\ref{v2i_res}}\;
    {Store $\mathcal{P}_t=\{(\boldsymbol{\mu}_t^m,\boldsymbol{\Sigma}_t^m)\}_{m\in\{v,t\}}$ using Eq.~\ref{new_distr}}\;
    \For{{each training mini-batch $\mathcal{B}\subset\mathcal{D}_{f,t}\cup\mathcal{D}_r$}}{
        \For{{each pair $(x^v,x^t)\in\mathcal{B}$}}{
            {Extract $(\mathbf{z}^v,\mathbf{z}^t)$ with CM-G using $\mathcal{C}_t$ via Eqs.~\ref{i2v}--\ref{v2i_res}}\;
            \For{{$i=1$ \KwTo $t$}}{
                {Compute $D_i^v(\mathbf{z}^v)$ and $D_i^t(\mathbf{z}^t)$ via Eqs.~\ref{eq1}--\ref{eq2}}\;
                {$d_i\leftarrow\tfrac12\bigl(D_i^v(\mathbf{z}^v)+D_i^t(\mathbf{z}^t)\bigr)$}\;
            }
            {Compute $\boldsymbol{\omega}=(\omega_1,\ldots,\omega_t)$ via Eqs.~\ref{ctg_closeness}--\ref{ctg_soft_weight}}\;
            {Encode $(x^v,x^t)$ with CM-G and weights $\boldsymbol{\omega}$, applying Eq.~\ref{sum_lora} at each LoRA insertion location; collect $(\mathbf{v},\mathbf{t})$}\;
        }
        {Compute $\mathcal{L}_{\mathrm{retain}}$ and $\mathcal{L}_{\mathrm{forget}}$ on the corresponding samples in $\mathcal{B}$ via Eqs.~\ref{retain_loss}--\ref{forget_loss}}\;
        {Update only $\mathrm{LoRA}_t$ by minimizing $\mathcal{L}_{\mathrm{retain}}-\lambda\mathcal{L}_{\mathrm{forget}}$ via Eq.~\ref{total_loss}}\;
    }
    {Freeze $\mathrm{LoRA}_t$}\;
}
{\Return $\{\mathrm{LoRA}_i,\mathcal{P}_i\}_{i=1}^{T}$}\;
\end{algorithm}

\begin{algorithm}[!t]
\small
\SetAlgoLined
\caption{{Inference procedure of CMCT-G for image-to-text retrieval}}
\label{alg:CMCT-G-infer}
\KwIn{{Image query $x^v$; complete text candidate set $\mathcal{T}=\{x_j^t\}_{j=1}^{N}$; frozen CLIP encoders $E_v^0,E_t^0$; task adapters $\{\mathrm{LoRA}_i\}_{i=1}^{T}$ and prototypes $\{\mathcal{P}_i\}_{i=1}^{T}$.}}
\KwOut{{Text candidates ranked by their matching scores for $x^v$.}}
\BlankLine
{$\mathbf{r}^v\leftarrow E_v^0(x^v)$}\;
{Compute $\{D_i^v(\mathbf{r}^v)\}_{i=1}^{T}$ via Eq.~\ref{eq1}}\;
\For{{$j=1$ \KwTo $N$}}{
  {$\mathbf{r}_j^t\leftarrow E_t^0(x_j^t)$}\;
  \For{{$i=1$ \KwTo $T$}}{
    {Compute $D_i^t(\mathbf{r}_j^t)$ via Eq.~\ref{eq2}}\;
    {$d_{ij}\leftarrow\tfrac12\bigl(D_i^v(\mathbf{r}^v)+D_i^t(\mathbf{r}_j^t)\bigr)$}\;
  }
  {Compute $\boldsymbol{\omega}_j$ via Eq.~\ref{eq:pair_routing}}\;
  {$\mathbf{z}_j^{\prime v}\leftarrow E_v(x^v;\{\omega_{ij}\mathrm{LoRA}_i\}_{i=1}^{T})$}\;
  {$\mathbf{z}_j^{\prime t}\leftarrow E_t(x_j^t;\{\omega_{ij}\mathrm{LoRA}_i\}_{i=1}^{T})$}\;
  {$s_j\leftarrow\cos(\mathbf{z}_j^{\prime v},\mathbf{z}_j^{\prime t})$}\;
}
{\Return candidates sorted by $s_j$ in descending order}\;
\end{algorithm}

\subsection{Cross-Task Guidance}
\label{Cross-Task_Guidance}
In CMCU, deletion requests for concepts such as \textit{[weapon]}, \textit{[blood]}, and \textit{[suicide]} arrive sequentially over time. Directly fine-tuning the entire cross-modal model for each new request would not only incur considerable computational and memory overhead, but also risk damaging the general knowledge obtained during large-scale pretraining. To enable efficient and localized parameter adaptation, {we adopt LoRA \cite{hu2022lora} and insert adapters of rank $r=16$ into \textit{k\_proj}, \textit{v\_proj}, \textit{out\_proj}, \textit{fc1}, and \textit{fc2} in every transformer block} of both the vision and text encoders. The projection matrix $\mathbf{W}$ is thus modified as:
\begin{equation}
\mathbf{W}'\mathbf{x} = \mathbf{W}\mathbf{x} + \Delta \mathbf{W}\mathbf{x} = \mathbf{W}\mathbf{x} + \mathbf{B}\mathbf{A}\mathbf{x},
\end{equation}
where $\mathbf{A} \in \mathbb{R}^{r \times d}$ and $\mathbf{B} \in \mathbb{R}^{d \times r}$ are low-rank learnable matrices, while $\mathbf{W}$ remains frozen during unlearning.

Although this parameter-efficient design allows concept-specific updates, assigning an independent LoRA adapter to every unlearning task soon becomes problematic. On the one hand, many target concepts exhibit strong semantic correlation or contextual co-occurrence (for example, \textit{[blood]} frequently appears with \textit{[suicide]}) \cite{zhou2022learning, Wang_2022_CVPR, smith2023coda}, which suggests that the knowledge learned from one task could benefit another if properly shared. On the other hand, continuously adding new LoRA adapters increases the risk of interference, as accumulated parameter shifts may begin to affect unrelated concepts. These observations motivate a more coordinated mechanism that allows related concepts to interact while keeping unrelated ones insulated.

To achieve this, we propose the \textbf{Cross-Task Guidance} module (\textbf{CT-G}), which leverages a prototype-based weighting mechanism to adaptively reuse LoRA modules across tasks. CT-G determines both the contribution of each accumulated LoRA adapter and the total strength of their combined update for the current sample. For each task, CT-G stores a covariance-aware prototype summary: the mean captures the task center, and the covariance captures feature dispersion for computing squared Mahalanobis routing distances. The task mean and covariance provide a compact summary for routing; the task features are not assumed to strictly follow a unimodal Gaussian distribution. We add diagonal regularization $\delta \mathbf{I}$ with $\delta=10^{-6}$ for numerical stability. For each unlearning task, a dedicated LoRA adapter is first initialized. \textcolor{black}{Before training task $t$, we extract the modality-specific features $\{\mathbf{z}_k^v\}_{k=1}^{n}$ and $\{\mathbf{z}_k^t\}_{k=1}^{n}$ from its forget set $\mathcal{D}_{f,t}$ with CM-G, following Eqs.~\ref{i2v}--\ref{v2i_res}. We then compute their empirical means and regularized covariances:}
{\color{black}
\begin{equation}
\label{new_distr}
\begin{aligned}
\boldsymbol{\mu}_t^m &= \frac{1}{n} \sum_{k=1}^{n} \mathbf{z}_k^m, \\
\boldsymbol{\Sigma}_t^m &= \frac{1}{n} \sum_{k=1}^{n} (\mathbf{z}_k^m - \boldsymbol{\mu}_t^m)(\mathbf{z}_k^m - \boldsymbol{\mu}_t^m)^\top + \delta \mathbf{I},
\end{aligned}
\end{equation}
}
where $m \in \{v, t\}$ denotes the modality; superscript $t$ denotes text, whereas task indices appear as subscripts. \textcolor{black}{Here, $n=|\mathcal{D}_{f,t}|$, and we store $\mathcal{P}_t=\{(\boldsymbol{\mu}_t^m,\boldsymbol{\Sigma}_t^m)\}_{m\in\{v,t\}}$ as the prototype summary for task $t$.}

{The task prototypes summarize features extracted with CM-G, which emphasizes the target concept. During inference, CT-G compares the original inputs' features from the frozen CLIP encoders with these prototypes in the shared image--text semantic space. Specifically, the image feature is compared with the text-side prototype, and the text feature with the vision-side prototype, to assess the relevance of the input pair to each deletion task.}

{For example, a deletion task may target the concept \textit{[weapon]}, while an associated image also contains a laptop and a table. The feature analysis in Section~\ref{sec:feature_analysis} shows that CM-G-guided features are more focused on the target concept. Prototypes constructed from these features therefore emphasize the weapon-related semantics. At inference, the image still contains all these objects; its feature and the candidate text feature are compared with the corresponding opposite-modality prototypes to determine the weight of this task's adapter.}

For a paired training example at task $t$, given embeddings
$(\mathbf{z}^v, \mathbf{z}^t)$, we compute its cross-modal squared
Mahalanobis distances to each stored task prototype
$i\in\{1,\ldots,t\}$ by referencing the opposite modality.
Because CLIP maps visual and textual representations into a
shared semantic space, each distance measures how one modality
relates to the opposite-modality prototype of the same deletion
task. Averaging the two directions gives a single bidirectional
cross-modal distance:
\begin{equation}
\label{eq1}
D_i^v(\mathbf{z}^v) = (\mathbf{z}^v - \boldsymbol{\mu}_i^t)^\top (\boldsymbol{\Sigma}_i^t)^{-1} (\mathbf{z}^v - \boldsymbol{\mu}_i^t),
\end{equation}
\begin{equation}
\label{eq2}
D_i^t(\mathbf{z}^t) = (\mathbf{z}^t - \boldsymbol{\mu}_i^v)^\top (\boldsymbol{\Sigma}_i^v)^{-1} (\mathbf{z}^t - \boldsymbol{\mu}_i^v),
\end{equation}
and aggregate the bidirectional distance as
\begin{equation}
\label{sum}
d_i = \frac{1}{2} \left(D_i^v + D_i^t\right).
\end{equation}
A smaller $d_i$ indicates stronger relevance to the $i$-th unlearning task prototype. Before LoRA aggregation, we convert this distance into a closeness score:
\begin{equation}
\label{ctg_closeness}
a_i = \exp(-d_i/\tau_r),
\end{equation}
where $\tau_r$ is the routing temperature and is set to $1$ by default in our experiments. For each $i\in\{1,\ldots,t\}$, the effective routing weight is
{
\begin{equation}
\label{ctg_soft_weight}
\omega_i =
\underbrace{
\max_{k \in \{1,\ldots,t\}} a_k
}_{\text{prototype-confidence gate}}
\cdot
\underbrace{
\frac{a_i}{\sum_{k=1}^{t}a_k}
}_{\text{relative adapter routing}}.
\end{equation}
}
The \underline{relative adapter routing} term distributes the adapter contribution according to prototype relevance. The shared \underline{prototype-confidence gate} controls its total magnitude because \textcolor{black}{$\sum_{i=1}^{t}\omega_i=\max_{1\leq k\leq t}a_k$}. A sample close to a stored forgetting prototype gives greater weight to the corresponding adapter. When the sample is far from every stored prototype, the total LoRA contribution decreases and the resulting representation approaches the frozen-backbone output.

The resulting weights aggregate the stored LoRA outputs:
\begin{equation}
\label{sum_lora}
\mathbf{o}' = \mathbf{W}\mathbf{x} + \sum_{i=1}^t \omega_i \cdot \text{LoRA}_i(\mathbf{x}) 
           = \mathbf{W}\mathbf{x} + \sum_{i=1}^t \omega_i \cdot \mathbf{B}_i \mathbf{A}_i \mathbf{x},
\end{equation}
where $\text{LoRA}_i(\mathbf{x})$ denotes the output of the $i$-th adapter. Soft cross-task weighting shares unlearning signals among related concepts while limiting adapter influence on retained samples far from the stored forgetting tasks.

\textcolor{black}{For each training image--text pair, the same routing weights are applied at every LoRA insertion location in both encoders. In Eq.~\ref{sum_lora}, $\mathbf{x}$ denotes the hidden state entering the corresponding layer.}

\paragraph{{Routing and ranking for retrieval.}}
\label{sec:retrieval_protocol}
{Following Safe-CLIP~\cite{poppi2024safe} and Hyperbolic Safe-CLIP~\cite{poppi2025hyperbolic}, we evaluate both image-to-text and text-to-image retrieval using single-modality queries. Conventional CLIP independently encodes the query and candidates and ranks them by cosine similarity. CMCT-G additionally computes routing weights for each query--candidate pair before obtaining the adapted embeddings and their matching score. We describe the two steps below using image-to-text retrieval.}

{\textit{Routing weights.} Given an image query $x^v$ and the complete text candidate set $\mathcal{T}=\{x_j^t\}_{j=1}^{N}$, the frozen CLIP encoders provide $\mathbf{r}^v=E_v^0(x^v)$ and $\mathbf{r}_j^t=E_t^0(x_j^t)$. Here, the superscript $0$ denotes the original encoders without CM-G or LoRA adapters. For each pair $(x^v,x_j^t)$, we substitute these embeddings into Eqs.~\ref{eq1}--\ref{eq2} to obtain $D_i^v(\mathbf{r}^v)$ and $D_i^t(\mathbf{r}_j^t)$ for each stored task prototype. Following Eqs.~\ref{sum}--\ref{ctg_soft_weight}, the pair's task distances and routing weights are}
{
\begin{equation}
\label{eq:pair_routing}
\begin{aligned}
d_{ij}&=\tfrac12\bigl(D_i^v(\mathbf{r}^v)+D_i^t(\mathbf{r}_j^t)\bigr),\\
a_{ij}&=\exp(-d_{ij}/\tau_r),\qquad
\omega_{ij}=\left(\max_{1\leq k\leq T}a_{kj}\right)
\frac{a_{ij}}{\sum_{k=1}^{T}a_{kj}}.
\end{aligned}
\end{equation}}
{Here, $T$ is the number of accumulated task adapters and $\tau_r=1$. The index $i$ identifies a task adapter, while $j$ identifies a candidate text; the maximum and sum are taken over task adapters. This gives a routing-weight vector $\boldsymbol{\omega}_j=(\omega_{1j},\ldots,\omega_{Tj})$ for each pair $(x^v,x_j^t)$.}

{\textit{Weighted encoding and ranking.} For each pair, we use $\boldsymbol{\omega}_j$ to weight the outputs of the task-specific LoRA adapters in both encoders through Eq.~\ref{sum_lora}. The same weights are used at every LoRA insertion location. The resulting visual and textual representations are}
{
\[
\begin{aligned}
\mathbf{z}_j^{\prime v}&=E_v\bigl(x^v;\{\omega_{ij}\mathrm{LoRA}_i\}_{i=1}^{T}\bigr),\\
\mathbf{z}_j^{\prime t}&=E_t\bigl(x_j^t;\{\omega_{ij}\mathrm{LoRA}_i\}_{i=1}^{T}\bigr).
\end{aligned}
\]
}
{Their cosine similarity gives the matching score:}
{
\begin{equation}
\label{eq:pair_score}
s_j=\frac{(\mathbf{z}_j^{\prime v})^\top\mathbf{z}_j^{\prime t}}
{\|\mathbf{z}_j^{\prime v}\|_2\,\|\mathbf{z}_j^{\prime t}\|_2}.
\end{equation}
}
{The same query image can have different adapted embeddings when paired with different texts, since the routing weights depend on the current pair. After computing $s_j$ for all $N$ candidates, we rank the texts in descending order of their scores. The ground-truth match is treated in the same way as every other candidate; its identity is used only to evaluate the resulting ranking.}

{Text-to-image retrieval follows the same procedure with the query and candidate modalities exchanged. Algorithm~\ref{alg:CMCT-G-infer} summarizes the image-to-text retrieval procedure.}

\subsection{Loss Function}
\label{sec:loss_function}

Following \cite{caitargeted}, we apply a contrastive loss on the retain set $\mathcal{D}_{r}$ to preserve the alignment of image–text pairs associated with concepts unrelated to the current unlearning target:
\begin{equation}
\label{retain_loss}
\mathcal{L}_{\text{retain}}(\theta, \mathcal{D}_{r}) = \frac{1}{2N_r} \sum_{i=1}^{N_r} \left( \ell_{\text{v-to-t}}(i) + \ell_{\text{t-to-v}}(i) \right),
\end{equation}
where
\begin{equation}
\begin{aligned}
\ell_{\text{v-to-t}}(i) &= - \log \frac{\exp(\text{cos}(\mathbf{v}_i, \mathbf{t}_i)/\tau_c)}{\sum_{j} \exp(\text{cos}(\mathbf{v}_i, \mathbf{t}_j)/\tau_c)} \\
\ell_{\text{t-to-v}}(i) &= - \log \frac{\exp(\text{cos}(\mathbf{t}_i, \mathbf{v}_i)/\tau_c)}{\sum_{j} \exp(\text{cos}(\mathbf{t}_i, \mathbf{v}_j)/\tau_c)}
\end{aligned}
\end{equation}
where $\tau_c$ denotes the contrastive-loss temperature.

To suppress alignment within the forget set $\mathcal{D}_{f,t}$, we introduce a cosine embedding loss that explicitly reduces the similarity between paired visual and textual representations:
\begin{equation}
\label{forget_loss}
\mathcal{L}_{\text{forget}}(\theta, \mathcal{D}_{f,t}) = \frac{1}{N_{f,t}} \sum_{i=1}^{N_{f,t}} \left( 1 - \text{cos}(\mathbf{v}_i, \mathbf{t}_i) \right).
\end{equation}

Here, $\mathbf{v}_i$ and $\mathbf{t}_i$ denote the visual and textual representations after sequential processing by CM-G and CT-G.

The final training objective is formulated as:
\begin{equation}
\label{total_loss}
\min_\theta \ \mathcal{L}_{\text{retain}}(\theta, \mathcal{D}_{r}) - \lambda \mathcal{L}_{\text{forget}}(\theta, \mathcal{D}_{f,t}),
\end{equation}
where $\lambda$ balances the strength of forgetting against the preservation of unrelated knowledge.

For partially overlapping forget/retain concepts, CMCT-G relies on the retain-side objective to preserve shared but non-target semantics. When a retained sample shares contextual information with a forgetting concept, the retain loss preserves its image--text alignment and discourages the update from suppressing all related semantics. For unrelated retained samples that are far from all stored prototypes, the prototype-confidence gate reduces the LoRA contribution. These mechanisms encourage the model to preserve contextual information that is not part of the deletion request.

For small forget sets, the forget loss is optimized jointly with the retain loss on CM-G-guided representations, while only the task-specific LoRA adapter is updated and the backbone remains frozen. These constraints reduce global representation drift, although the method may still be sensitive to extremely small forget sets.

\section{Experiments}

\subsection{Semantic-Centric Dataset Construction}

To evaluate methods under the CMCU setting, we construct two \emph{semantic-centric} image–text datasets: (i) an \textbf{unsafe-level} concept dataset, \textbf{UnSafeBench}, and (ii) an \textbf{object-level} concept dataset, \textbf{ObjectBench}. Both datasets are partitioned by a set of predefined concepts. Each concept is represented by paired image–text samples containing multiple co-occurring elements and weakly aligned cross-modal content. This design deviates from the \textit{object-centric} assumptions commonly adopted in previous unlearning benchmarks.

Based on these two datasets, we construct four evaluation benchmarks by instantiating both \emph{single-concept} and \emph{grouped-concept} continual unlearning protocols, following \cite{zhou2025revisiting, feng2025fg, lu2024mace}.

\paragraph{UnSafeBench (unsafe-level concepts).}
Following \cite{poppi2024safe}, we define 20 unsafe concepts (i.e., \textit{hate}, \textit{harassment}, \textit{violence}, \textit{suffering}, \textit{humiliation}, \textit{harm}, \textit{suicide}, \textit{sexual}, \textit{nudity}, \textit{bodily fluids}, \textit{blood}, \textit{obscene gestures}, \textit{illegal activity}, \textit{drug use}, \textit{theft}, \textit{vandalism}, \textit{weapons}, \textit{abuse}, \textit{brutality}, \textit{cruelty}) and generate corresponding multimodal pairs.
For each concept, we first employ the {LLaMA-2}~\cite{touvron2023llama} model fine-tuned with Direct Preference Optimization (DPO)~\cite{rafailov2023direct} to produce descriptive texts\footnote{Please refer to \cite{poppi2024safe} for the detailed fine-tuning process.}. We then use an open-source diffusion model on Hugging Face specifically designed for generating unsafe images\footnote{\url{https://huggingface.co/stablediffusionapi/newrealityxl-global-nsfw}} to synthesize images conditioned on the generated texts, forming candidate image--text pairs for each unsafe concept. During construction, we observed that direct image generation may produce low-quality samples, such as visually distorted images, missing target visual content, or weak concept-image correspondence. Therefore, we use a CLIP-score filter to select generated images with higher concept--image similarity: given the prompt ``\textit{a sentence about [unsafe concept]}'', we compute the cosine similarity between the CLIP text embedding of this prompt and the CLIP image embedding of each generated image, retaining image--text pairs with a similarity score greater than 0.6. Because this pipeline is synthetic, \textbf{UnSafeBench} is positioned as a controlled semantic-centric stress test rather than a complete substitute for real unsafe content; \textbf{ObjectBench} provides complementary evaluation with natural image--caption pairs.

Based on this dataset, we build two benchmarks:  
(1) \textit{UnSafeBench-Single20}, which consists of 20 sequential tasks, each removing one unsafe concept; and  
(2) \textit{UnSafeBench-Group7}, which merges semantically related concepts into seven clusters, mimicking coarse-grained policy-level removals. The grouping follows the taxonomy in \cite{schramowski2023safe} and is summarized in Table~\ref{tab:unsafe_group7}.

\begin{table}[t]
\centering
\resizebox{\columnwidth}{!}{
\begin{tabular}{p{0.58\columnwidth}<{\raggedright} p{0.35\columnwidth}}
\toprule
\textbf{Single Concepts} & \textbf{Grouped Concepts} \\
\midrule
\textit{Hate} & \textit{Hate} \\
\rowcolor{gray!30}
\textit{Harassment} & \textit{Harassment} \\
\textit{Violence, Suffering, Humiliation, Harm, Abuse, Brutality, Cruelty} & \multirow{2}{*}{\textit{Violence}} \\
\rowcolor{gray!30}
\textit{Suicide} & \textit{Self-harm} \\
\textit{Sexual, Nudity} & \textit{Sexual} \\
\rowcolor{gray!30}
\textit{Bodily Fluids, Blood, Obscene Gestures} & \multirow{2}{*}{\textit{Shocking}} \\

\textit{Illegal Activity, Drug Use, Theft, Vandalism, Weapons} & \multirow{2}{*}{\textit{Illegal activity}} \\
\bottomrule
\end{tabular}
}
\caption{\textbf{Mapping between single and grouped concepts in {UnSafeBench}}.
Each grouped category consolidates semantically related unsafe concepts following the taxonomy of~\cite{schramowski2023safe}.}
\label{tab:unsafe_group7}
\end{table}

\paragraph{ObjectBench (object-level concepts).}
We construct \textbf{ObjectBench} from the MS-COCO~\cite{lin2014microsoft} dataset by reorganizing its original image–caption pairs and instance annotations according to the predefined \textbf{categories}.
Each object category in COCO serves as an object-level concept, resulting in 80 fine-grained concepts (e.g., \textit{person}, \textit{car}, \textit{dog}, \textit{chair}, \textit{pizza}).
Following the official taxonomy, these 80 object-level concepts are further grouped into 12 higher-level \textbf{supercategories} (e.g., \textit{vehicle}, \textit{animal}, \textit{furniture}, \textit{food}, \textit{sports}).
This hierarchical organization allows us to define both single concept and grouped concept continual unlearning protocols.
The correspondence between COCO supercategories and categories is summarized in Table~\ref{tab:coco_supercat_swapped}.

Based on this dataset, we build two benchmarks: (1) \textit{ObjectBench-Single20}, which selects 20 representative target concepts from the 80 categories as unlearning targets. Each concept is sequentially unlearned, while the remaining 60 categories serve as preserved unrelated concepts. This benchmark evaluates fine-grained object-level forgetting and knowledge retention in continual unlearning. (2) \textit{ObjectBench-Group5}, which selects 5 supercategories from the 12 COCO supercategories as unlearning targets, while the remaining 7 supercategories serve as preserved unrelated concepts. This benchmark emphasizes higher-level semantic forgetting and evaluates group-wise unlearning performance.

\begin{table}[t]
\centering
\resizebox{\columnwidth}{!}{
\begin{tabular}{>{\raggedright\arraybackslash}p{0.6\columnwidth}
                >{\raggedright\arraybackslash}p{0.3\columnwidth}}
\toprule
\textbf{Categories} & \textbf{Supercategories} \\
\midrule
\textit{Person} & \textit{Person} \\
\rowcolor{gray!30}
\textit{Bicycle}, \textit{Car}, \textit{Motorcycle}, \textit{Airplane}, \textit{Bus}, \textit{Train}, \textit{Truck}, \textit{Boat} & \multirow{2}{*}{\textit{Vehicle}} \\
\textit{Traffic Light}, \textit{Fire Hydrant}, \textit{Stop Sign}, \textit{Bench} & \multirow{2}{*}{\textit{Outdoor}} \\
\rowcolor{gray!30}
\textit{Bird}, \textit{Cat}, \textit{Dog}, \textit{Horse}, \textit{Sheep}, \textit{Cow}, \textit{Elephant}, \textit{Bear}, \textit{Zebra}, \textit{Giraffe} & \multirow{2}{*}{\textit{Animal}} \\

\textit{Backpack}, \textit{Umbrella}, \textit{Handbag}, \textit{Tie}, \textit{Suitcase} & \multirow{2}{*}{\textit{Accessory}} \\
\rowcolor{gray!30}
\textit{Frisbee}, \textit{Skis}, \textit{Snowboard}, \textit{Sports Ball}, \textit{Kite}, \textit{Baseball Bat}, \textit{Baseball Glove}, \textit{Skateboard}, \textit{Surfboard}, \textit{Tennis Racket} & \multirow{3}{*}{\textit{Sports}} \\

\textit{Bottle}, \textit{Wine Glass}, \textit{Cup}, \textit{Fork}, \textit{Knife}, \textit{Spoon}, \textit{Bowl} & \multirow{2}{*}{\textit{Kitchen}} \\
\rowcolor{gray!30}
\textit{Banana}, \textit{Apple}, \textit{Sandwich}, \textit{Orange}, \textit{Broccoli}, \textit{Carrot}, \textit{Hot Dog}, \textit{Pizza}, \textit{Donut}, \textit{Cake} & \multirow{3}{*}{\textit{Food}} \\

\textit{Chair}, \textit{Couch}, \textit{Potted Plant}, \textit{Bed}, \textit{Dining Table}, \textit{Toilet} & \multirow{2}{*}{\textit{Furniture}} \\
\rowcolor{gray!30}
\textit{TV}, \textit{Laptop}, \textit{Mouse}, \textit{Remote}, \textit{Keyboard}, \textit{Cell Phone} & \multirow{2}{*}{\textit{Electronic}} \\

\textit{Microwave}, \textit{Oven}, \textit{Toaster}, \textit{Sink}, \textit{Refrigerator} & \multirow{2}{*}{\textit{Appliance}} \\
\rowcolor{gray!30}
\textit{Book}, \textit{Clock}, \textit{Vase}, \textit{Scissors}, \textit{Teddy Bear}, \textit{Hair Drier}, \textit{Toothbrush} & \multirow{2}{*}{\textit{Indoor}} \\
\bottomrule
\end{tabular}
}
\caption{\textbf{Mapping between categories and supercategories in ObjectBench.} Each supercategory groups multiple fine-grained categories that serve as atomic concepts in \textbf{ObjectBench}.}
\label{tab:coco_supercat_swapped}
\end{table}

Overall, the two benchmarks provide complementary semantic-centric evaluation settings. \textbf{UnSafeBench} follows the synthetic unsafe-data construction practice used in Safe-CLIP~\cite{poppi2024safe}, and further reorganizes the controlled unsafe image-text pairs into concept-level and group-level CMCU protocols with explicit forget/retain splits. \textbf{ObjectBench} serves as a real-image counterpart built from MS-COCO image-caption pairs and instance annotations. Its semantic-centric property comes from natural COCO scenes, where each image usually contains multiple co-occurring objects, the target concept may occupy only a small part of the scene, and the caption may mention only a subset of the visual content. Therefore, unlike conventional object-centric unlearning benchmarks that assume each sample is dominated by a single class, \textbf{ObjectBench} evaluates whether continual unlearning methods remain effective when the target concept is embedded in natural multi-object image-text pairs.

\paragraph{{Benchmark validation and characterization.}}
\label{sec:benchmark_validation}
{To quantitatively validate and characterize the two benchmarks, we sample 400 image--text pairs from each: 20 pairs per concept across the 20 unsafe concepts in Table~\ref{tab:unsafe_group7}, and five pairs per category across the 80 ObjectBench categories in Table~\ref{tab:coco_supercat_swapped}. Two human annotators and two multimodal models, GPT-5.6 Sol~\cite{openai2026gpt56sol} and Kimi-K3~\cite{moonshot2026kimik3}, independently assess concept presence, additional visual and textual content, and image--caption content mismatch. Appendices~\ref{sec:qualitative_validation_results} and~\ref{sec:validation_examples} provide the quantitative results, validation protocol, and prompt template, together with two illustrative examples, one from each benchmark, showing the annotations from all four evaluators.}

\begin{table*}[t!]
\centering
\resizebox{\textwidth}{!}{
\begin{tabular}{lllllllllll}
\toprule
\multirow{3}{*}{Methods} & \multirow{3}{*}{Reference} &\multirow{3}{*}{\begin{tabular}[c]{@{}c@{}}Tunable\\ Ratio $\downarrow$ \end{tabular}} & \multicolumn{4}{c}{\textit{UnSafeBench-Single20}} & \multicolumn{4}{c}{\textit{UnSafeBench-Group7}} \\ 
\cmidrule(r){4-7} \cmidrule(r){8-11} 
 & & & \multicolumn{2}{c}{$\mathcal{F}_B \downarrow$} & \multicolumn{2}{c}{$\mathcal{R}_B \uparrow$} & \multicolumn{2}{c}{$\mathcal{F}_B \downarrow$} & \multicolumn{2}{c}{$\mathcal{R}_B \uparrow$} \\
 \cmidrule(r){4-5} \cmidrule(r){6-7} \cmidrule(r){8-9} \cmidrule(r){10-11} 
 & & & \textbf{T}-to-\textbf{V} & \textbf{V}-to-\textbf{T} & \textbf{T}-to-\textbf{V} & \textbf{V}-to-\textbf{T} & \textbf{T}-to-\textbf{V} & \textbf{V}-to-\textbf{T} & \textbf{T}-to-\textbf{V} & \textbf{V}-to-\textbf{T} \\ \midrule
Pretrained & - & - & 95.20 & 95.20 & 66.60 & 64.30 & 92.58 & 89.30 & 56.20 & 56.10 \\
\midrule
\multicolumn{5}{l}{\textcolor{orange}{\textit{(a) continual learning methods}}} \\
EWC* & PNAS'17 & 100\% & 38.17 & 29.04 & 19.23 & 52.04  & 29.30 & 23.03 & 43.19 & 47.38 \\
MAS* & ECCV'18 & 100\% & 28.90 & 18.33 & 23.17 & 49.30  & 30.18 & \underline{17.04} & 32.24 & 44.59 \\
FDR & ICLR'19 & 100\% & 29.36 & 21.02 & 19.36 & 19.17  & 24.33 & 18.10 & 39.30 & 19.04 \\
\midrule
\multicolumn{5}{l}{\textcolor{orange}{\textit{(b) machine unlearning methods}}} \\
Retrain & - & 100\% & \underline{20.91} & \underline{17.39} & 23.05 & 10.24 & \underline{19.20} & 23.93 & 32.63 & 20.03 \\
SCRUB & Neurips'23 & 100\% & 32.17 & 28.15 & 42.70 & 21.06  & 29.52 & 36.90 & 41.97 & 45.19 \\
SalUn & ICLR'24 & 100\% & 45.23 & 42.09 & 46.13 & 31.20  & 29.03 & 23.18 & \underline{50.02} & 42.32 \\
SLUG & ICML'25 & 1.00\% & 39.14 & 33.11 & 49.30 & 30.19  & 27.19 & 19.07 & 38.26 & 37.11 \\
CLIP-ERASE & ACL'25 & 100\% & 26.90 & 19.30 & 56.33 & 45.36  & 39.38 & 22.04 & 39.72 & 49.19 \\
Safe-CLIP & ECCV'24 & 1.71\% & 61.10 & 52.10 & 58.60 & 58.10 & 38.27 & 19.55 & 38.23 & 50.20 \\
Hyperbolic Safe-CLIP & CVPR'25 & 1.71\% & 28.40 & 21.85 & \underline{59.20} & \underline{58.70} & 34.80 & 18.40 & 45.70 & \underline{51.10} \\
\midrule
\multicolumn{5}{l}{\textcolor{orange}{\textit{(c) continual unlearning methods}}} \\
GS-LoRA & CVPR'24 & 1.71\% & 40.33 & 33.95 & 53.20 & 49.88  & 39.33 & 43.22 & 42.74 & 45.10 \\
FG-OrIU & ICCV'25 & 1.71\% & 35.97 & 29.14 & 56.37 & 51.23  & 37.20 & 39.11 & 45.26 & 47.23 \\
\midrule
\rowcolor{gray!30}
\textbf{CMCT-G} & - & 1.71\% & \textbf{17.50} & \textbf{9.50} & \textbf{62.40} & \textbf{60.30}  & \textbf{16.70} & \textbf{8.33} & \textbf{53.24} & \textbf{52.37} \\
\bottomrule
\end{tabular}%
}
\caption{
CMCU results on the \textit{UnSafeBench-Single20} and \textit{UnSafeBench-Group7} benchmarks. 
\textbf{T}-to-\textbf{V} and \textbf{V}-to-\textbf{T} denote text-to-vision and vision-to-text retrieval, respectively. 
$^*$ indicates original methods augmented with a rehearsal buffer. 
“Retrain” refers to models retrained using replay data, with training epochs kept identical to other methods for fair comparison. 
“Pretrained” denotes results before any unlearning. 
The best results are highlighted in \textbf{bold}, and the second-best results are \underline{underlined}.}
\label{tab:single-recog-caisa100_main}
\end{table*}

\begin{table*}[t!]
\centering
\resizebox{\textwidth}{!}{
\begin{tabular}{lllllllllll}
\toprule
\multirow{3}{*}{Methods} & \multirow{3}{*}{Reference} &\multirow{3}{*}{\begin{tabular}[c]{@{}c@{}}Tunable\\ Ratio $\downarrow$ \end{tabular}} & \multicolumn{4}{c}{\textit{ObjectBench-Single20}} & \multicolumn{4}{c}{\textit{ObjectBench-Group5}} \\ 
\cmidrule(r){4-7} \cmidrule(r){8-11} 
 & & & \multicolumn{2}{c}{$\mathcal{F}_B \downarrow$} & \multicolumn{2}{c}{$\mathcal{R}_B \uparrow$} & \multicolumn{2}{c}{$\mathcal{F}_B \downarrow$} & \multicolumn{2}{c}{$\mathcal{R}_B \uparrow$} \\
 \cmidrule(r){4-5} \cmidrule(r){6-7} \cmidrule(r){8-9} \cmidrule(r){10-11} 
 & & & \textbf{T}-to-\textbf{V} & \textbf{V}-to-\textbf{T} & \textbf{T}-to-\textbf{V} & \textbf{V}-to-\textbf{T} & \textbf{T}-to-\textbf{V} & \textbf{V}-to-\textbf{T} & \textbf{T}-to-\textbf{V} & \textbf{V}-to-\textbf{T} \\ \midrule
Pretrained & - & - & 98.20 & 97.10 & 79.32 & 78.50      & 95.43 & 92.15 & 69.31 & 79.62 \\
\midrule
\multicolumn{5}{l}{\textcolor{orange}{\textit{(a) continual learning methods}}} \\
EWC* & PNAS'17 & 100\% & 39.01 & 42.22 & 45.98 & 54.23 & 56.71 & 37.06 & 42.17 & 64.20 \\
MAS* & ECCV'18 & 100\% & 49.21 & 39.17 & 54.10 & 47.39 & 63.12 & 58.13 & 36.15 & 56.50 \\
FDR & ICLR'19 & 100\% & 66.70 & 42.33 & 32.11 & 29.38 & 61.25 & 39.15 & 38.20 & 27.31 \\
\midrule
\multicolumn{5}{l}{\textcolor{orange}{\textit{(b) machine unlearning methods}}} \\
Retrain & - & 100\% & 33.80 & 29.23 & 19.03 & 39.54 & 28.43 & 28.33 & 42.11 & 39.14 \\
SCRUB & NeurIPS'23 & 100\% & 34.19 & 39.17 & 47.60 & 34.04 & 30.60 & 40.56 & 46.35 & 37.29 \\
SalUn & ICLR'24 & 100\% & 33.57 & 47.86 & 44.23 & 35.42 & 38.29 & 37.82 & 49.26 & 48.30 \\
SLUG & ICML'25 & 1.00\% & 39.21 & 45.32 & 45.78 & 33.05 & 29.10 & 29.03 & 37.18 & 39.06 \\
CLIP-ERASE & ACL'25 & 100\% & 46.20 & 29.34 & 52.04 & 41.93 & 47.27 & 30.08 & 35.34 & 65.35 \\
Safe-CLIP & ECCV'24 & 1.71\% & 45.98 & 21.00 & 57.90 & 51.08 & 47.37 & 41.92 & 47.92 & 69.48 \\
Hyperbolic Safe-CLIP & CVPR'25 & 1.71\% & 40.72 & \underline{19.80} & 59.40 & 54.70 & 42.10 & 35.20 & 56.20 & \underline{70.10} \\
\midrule
\multicolumn{5}{l}{\textcolor{orange}{\textit{(c) continual unlearning methods}}} \\
GS-LoRA & CVPR'24 & 1.71\% & 44.37 & 24.11 & 55.23 & 49.36 & \underline{28.31} & \underline{22.10} & 59.72 & 65.70 \\
FG-OrIU & ICCV'25 & 1.71\% & \underline{31.20} & 20.30 & \underline{61.23} & \underline{55.24} & 36.14 & 28.29 & \underline{60.21} & 68.24 \\
\midrule
\rowcolor{gray!30}
\textbf{CMCT-G} & - & 1.71\% & \textbf{21.29} & \textbf{11.34} & \textbf{68.37} & \textbf{64.29} & \textbf{20.33} & \textbf{12.45} & \textbf{63.24} & \textbf{73.19} \\
\bottomrule
\end{tabular}
}
\caption{
CMCU results on the \textit{ObjectBench-Single20} and \textit{ObjectBench-Group5} benchmarks. The best results are highlighted in \textbf{bold}, and the second-best results are \underline{underlined}.}
\label{tab:objectbench-results}
\end{table*}

\subsection{Experimental Setup}

\paragraph{Training Details} All experiments are implemented using PyTorch and run on 8 NVIDIA A800 GPUs with Distributed Data Parallel. Following prior work \cite{poppi2024safe, poppi2025hyperbolic, caitargeted}, we use \textbf{CLIP-ViT-Large/14} as the backbone. For all baseline methods, we strictly re-implement their official code under the CMCU task to ensure fair comparison. During unlearning, we train the model for 10 epochs with a per-GPU batch size of 32. The learning rate is set to 1e-3, and the rank of the LoRA adapters is fixed at 16. All baselines use the same forget/retain splits, backbone, training schedule, batch size, and LoRA rank whenever applicable. Since most baselines do not define an internal mechanism for using localized visual masks, their original input protocols are kept unchanged in the main comparison. Following the evaluation protocol in APER~\cite{zhou2025revisiting}, the main results are averaged over three independent runs with different random seeds: 37, 1337, and 13337. We additionally evaluate task-order sensitivity on \textit{UnSafeBench-Single20} and \textit{ObjectBench-Single20}, as summarized in Table~\ref{tab:order_variance}.

\paragraph{Evaluation Metric} 
Following \cite{zhou2025revisiting, poppi2024safe}, we denote cross-modal retrieval accuracy on all previously forgotten concepts after unlearning the $b$-th concept as $\mathcal{F}_b$, and that on the {retain set} as $\mathcal{R}_b$. We report $\mathcal{F}_B$ (final retrieval performance on all forgotten concepts) and $\mathcal{R}_B$ (final retrieval performance on the {retain set}) as evaluation metrics. Lower $\mathcal{F}_B$ indicates that the unlearned model consistently suppresses retrieval for all previously unlearned target concepts, achieving effective continual unlearning. Higher $\mathcal{R}_B$ reflects the unlearned model’s robustness in preserving retrieval performance on unrelated concepts.
These retrieval-based metrics are aligned with the training objective and deployment usage of CLIP-like models, whose image and text encoders are optimized through cross-modal semantic alignment.

\begin{table}[t]
\centering
\resizebox{\linewidth}{!}{
\begin{tabular}{llcccc}
\toprule
\multirow{2}{*}{{Benchmark}} & \multirow{2}{*}{{Task Order}} & \multicolumn{2}{c}{\textbf{$\mathcal{F}_B \downarrow$}} & \multicolumn{2}{c}{\textbf{$\mathcal{R}_B \uparrow$}} \\
\cmidrule(lr){3-4} \cmidrule(lr){5-6}
& & \textbf{T-to-V} & \textbf{V-to-T} & \textbf{T-to-V} & \textbf{V-to-T} \\
\midrule
\multirow{3}{*}{\makecell{\textit{UnSafeBench}\\\textit{-Single20}}}
& Original & 17.50 & 9.50 & 62.40 & 60.30 \\
& Reverse & 16.23 & 8.36 & 63.17 & 59.73 \\
& Random & 18.64 & 10.09 & 61.40 & 58.08 \\
\midrule
\multirow{3}{*}{\makecell{\textit{ObjectBench}\\\textit{-Single20}}}
& Original & 21.29 & 11.34 & 68.37 & 64.29 \\
& Reverse & 22.17 & 12.83 & 68.08 & 63.40 \\
& Random & 22.09 & 11.79 & 67.91 & 63.80 \\
\bottomrule
\end{tabular}
}
\caption{Task-order sensitivity analysis using the same bidirectional forgetting and retention metrics as the main evaluation.}
\label{tab:order_variance}
\end{table}

\begin{table}[t]
\centering
\resizebox{\linewidth}{!}{
\begin{tabular}{lcc}
\toprule
{Setting} & {\makecell{LoRA Params/\\Persistent Storage}} & {\makecell{Train Time\\ /Task/Epoch}} \\
\midrule
CMCT-G-1task  & 8.65M / 33.0 MiB & 3.58 min \\
CMCT-G-5task  & 43.25M / 165.0 MiB & 3.58 min \\
CMCT-G-10task & 86.51M / 330.0 MiB & 3.58 min \\
\rowcolor{gray!30}
\textbf{CMCT-G-20task} & \textbf{173.02M / 660.0 MiB} & \textbf{3.58 min} \\
\bottomrule
\end{tabular}
}
\captionsetup{justification=raggedright,singlelinecheck=false}
\caption{Efficiency and scalability analysis on \textit{UnSafeBench-Single20}. Storage denotes persistent LoRA adapter storage in FP32, and training time is reported per task per epoch.}
\label{tab:efficiency}
\end{table}

\subsection{Quantitative Analysis}
We compare \textbf{CMCT-G} with three categories of baselines: \textit{(a) continual learning (CL) methods}, including EWC~\cite{kirkpatrick2017overcoming}, MAS~\cite{aljundi2018memory}, and FDR~\cite{benjamin2018measuring}; \textit{(b) machine unlearning (MU) methods}, including SCRUB~\cite{kurmanji2023towards}, SalUn~\cite{fan2023salun}, SLUG~\cite{caitargeted}, CLIP-ERASE~\cite{yang2024cliperase}, Safe-CLIP~\cite{poppi2024safe}, and Hyperbolic Safe-CLIP~\cite{poppi2025hyperbolic}; and \textit{(c) continual unlearning (CU) methods}, including GS-LoRA~\cite{zhao2024continual} and FG-OrIU~\cite{feng2025fg}. Results on the CMCU task are presented in Table~\ref{tab:single-recog-caisa100_main} and Table~\ref{tab:objectbench-results}. \textbf{CMCT-G consistently outperforms all baseline methods across \textit{UnSafeBench-Single20, UnSafeBench-Group7, ObjectBench-Single20, and ObjectBench-Group5}.} Specifically, CL methods do not provide sufficient concept-level forgetting, as exemplified by FDR, which yields high $\mathcal{F}_B$ in \textbf{T}-to-\textbf{V} retrieval on both benchmarks. Existing MU and CU baselines are also limited by their original design assumptions on object-centric or static unlearning settings. When applied to semantic-centric image--text pairs, unrelated content and mismatched cross-modal elements can cause imprecise forgetting or retention degradation. For instance, SCRUB substantially reduces $\mathcal{R}_B$, indicating compromised retrieval utility on unrelated concepts. In contrast, CMCT-G achieves a stronger forgetting-retention balance by combining cross-modality guidance with task-aware adapter routing.

\begin{figure*}[t]
    \centering
    \includegraphics[width=\textwidth]{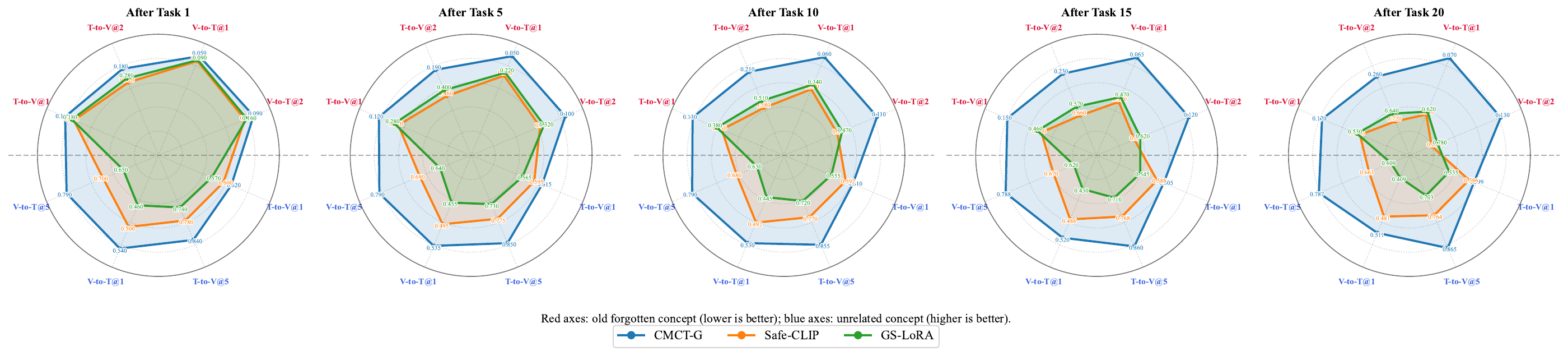}
    \caption{Trajectory-level analysis on \textit{UnSafeBench-Single20} across continual unlearning tasks. Unlike the benchmark-level aggregate results in the main tables, this figure tracks the first deleted concept on the red axes and unrelated retained concepts on the blue axes at the same checkpoints. We report retrieval performance after the $1$-st, $5$-th, $10$-th, $15$-th, and $20$-th unlearning tasks.}
    \label{fig:task_trajectory}
\end{figure*}

Figure~\ref{fig:task_trajectory} further expands the final-state summary in Figure~\ref{fig:radar}(b) into a task-wise trajectory analysis. Unlike the benchmark-level aggregate $\mathcal{F}_B/\mathcal{R}_B$ results in the main tables, this figure focuses on a representative early-forgotten concept and the unrelated retained subset across the continual unlearning process. CMCT-G keeps the early forgotten concept at a consistently low retrieval level after later deletion tasks, suggesting that the concept is not noticeably revived. At the same time, retrieval on unrelated concepts remains relatively stable. In comparison, Safe-CLIP and GS-LoRA show larger rebound on the early-forgotten concept or larger fluctuation in retained performance at the same checkpoints.

Table~\ref{tab:efficiency} reports the storage and training cost of CMCT-G. The adapter for one task has 8.65M parameters and requires 33.0 MiB of persistent storage in FP32. For $T$ deletion tasks, persistent LoRA storage is approximately $33.0T$ MiB in FP32, while full prototype routing and LoRA aggregation scale linearly with $T$. The accumulated adapters therefore require 33.0, 165.0, 330.0, and 660.0 MiB after 1, 5, 10, and 20 tasks, respectively. Previously learned adapters are retained for routing and aggregation but remain frozen; only the current adapter requires gradients. No noticeable increase in per-task per-epoch training time is observed from 1 to 20 tasks; the measured value remains 3.58 min. Grounded-SAM is performed as offline preprocessing for new deletion requests and costs 0.42 s/img on average; it is not invoked for inference queries.

\begin{figure*}
    \centering
    \includegraphics[width=1\textwidth]{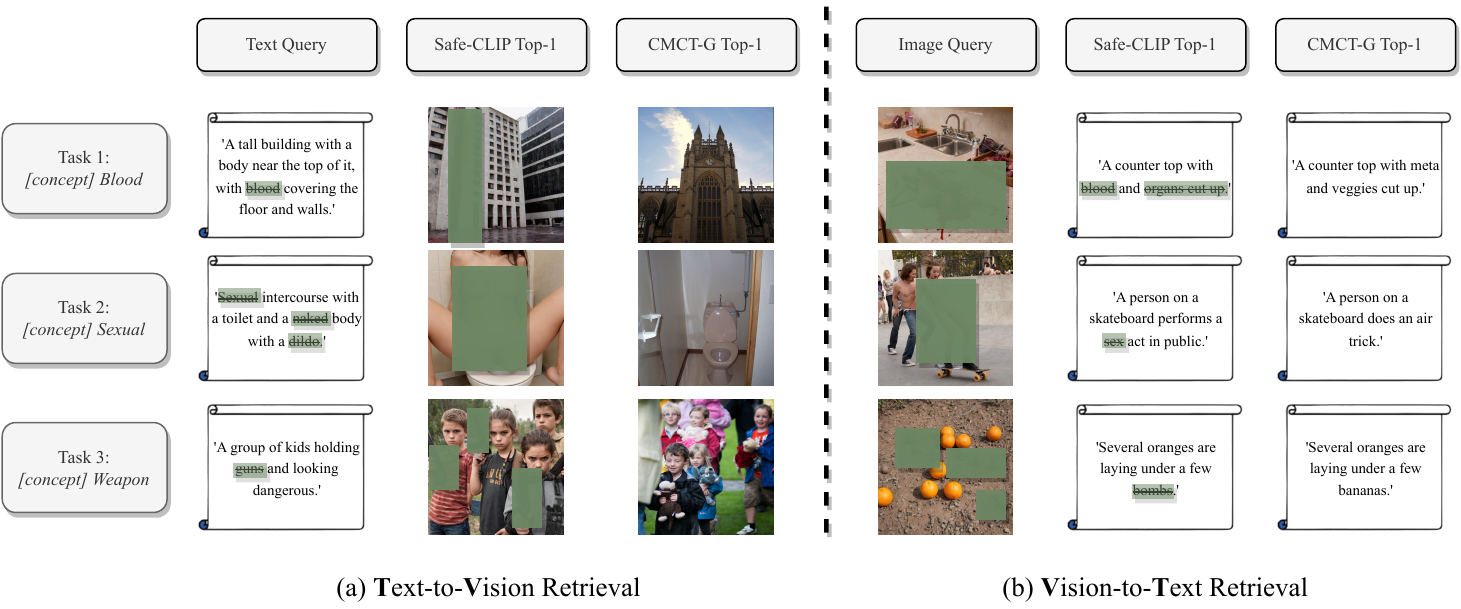}
    \caption{
    Top-1 retrieval results for images (a) and text (b) using the unlearned CLIP models obtained by Safe-CLIP and our CMCT-G, where both text and image queries are evaluated. 
    We select three representative concepts from \textit{UnSafeBench-Single20}, namely \textit{[Blood]}, \textit{[Sexual]}, and \textit{[Weapon]}, as illustrative examples. 
    Due to safety and compliance requirements, the images are partially obscured with a semi-transparent dark green mask at 95\% opacity. Similarly, the text results are masked, with sensitive words redacted using strikethrough formatting. These overlays are applied only to the displayed figure; all retrieval experiments and quantitative results use the original, unmasked samples.
    }

    \label{fig:Qualitative}
\end{figure*}

\subsection{Qualitative Comparison}

The CMCU task presents two major issues: achieving precise unlearning under semantic-centric datasets, and maintaining effective and robust performance in continual unlearning when handling sequential deletion requests. These issues have significantly limited the effectiveness of previous methods. As shown in Figure~\ref{fig:Qualitative}, our CMCT-G addresses these issues more successfully than Safe-CLIP. It can retrieve safe images given unsafe text queries and retrieve safe sentences given unsafe image queries.

\begin{table}[t!]
  \centering
    \small
  \resizebox{\columnwidth}{!}{
  \begin{tabular}{lllll}
  \toprule
    \multirow{2}{*}{Variants} & \multicolumn{2}{c}{{$\mathcal{F}_B \downarrow$}} & \multicolumn{2}{c}{$\mathcal{R}_B \uparrow$} \\
    \cmidrule(lr){2-3} \cmidrule(lr){4-5} 
    & {\textbf{T}-to-\textbf{V}} & {\textbf{V}-to-\textbf{T}} & {\textbf{T}-to-\textbf{V}} & {\textbf{V}-to-\textbf{T}} \\
    \midrule 
    Pre-train  & 95.20  & 95.20 & 66.60 & 64.30 \\
    \rowcolor{gray!30}
    \textbf{CMCT-G}  & \textbf{17.50}  & \textbf{9.50} & \textbf{62.40} & \textbf{60.30} \\
    \quad w/o \textit{CM-G-text}  & 26.33  & 23.05 & 56.70 & 50.80 \\
    \quad w/o \textit{CM-G-image}  & 35.20  & 27.34 & 55.43 & 53.17 \\
    \quad w/o \textit{CT-G}  & 28.17  & 26.19 & 52.10 & 52.03 \\
  \bottomrule
\end{tabular}
}
\caption{Module ablation studies on \textit{UnSafeBench-Single20}. We remove each component to evaluate its contribution to CMCT-G.}
\label{tab:ablation_table}
\end{table}

\begin{table}[t]
\centering
\resizebox{\linewidth}{!}{
\begin{tabular}{llcccc}
\toprule
\multirow{2}{*}{{Setting}} & \multirow{2}{*}{{Visual Cue}} & \multicolumn{2}{c}{\textbf{$\mathcal{F}_B \downarrow$}} & \multicolumn{2}{c}{\textbf{$\mathcal{R}_B \uparrow$}} \\
\cmidrule(lr){3-4}\cmidrule(lr){5-6}
& & \textbf{T-to-V} & \textbf{V-to-T} & \textbf{T-to-V} & \textbf{V-to-T} \\
\midrule
\rowcolor{gray!30}
\textbf{CMCT-G} & \textbf{Original} & \textbf{17.50} & \textbf{9.50} & \textbf{62.40} & \textbf{60.30} \\
Noisy 10\% & Perturbed & 18.64 & 10.51 & 61.91 & 59.61 \\
Noisy 30\% & Perturbed & 22.47 & 12.33 & 60.38 & 57.77 \\
Noisy 50\% & Perturbed & 26.04 & 15.98 & 58.52 & 56.10 \\
\bottomrule
\end{tabular}
}
\caption{Sensitivity analysis to noisy Grounded-SAM masks on \textit{UnSafeBench-Single20}. We perturb the original Grounded-SAM mask at a ratio of $\epsilon$ and use the perturbed mask to obtain the concept-purified visual cue for CM-G.}
\label{tab:cue_robustness}
\end{table}

\begin{figure*}[t]
  \centering
  \includegraphics[width=0.92\textwidth]{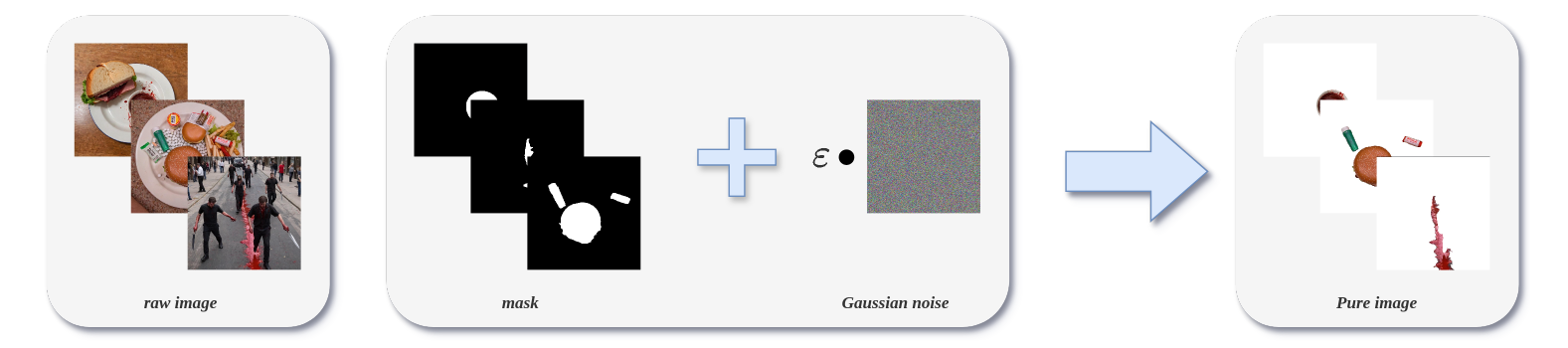}
  \caption{Qualitative illustration of the noisy-grounding sensitivity analysis. Starting from the raw image and the Grounded-SAM mask, we simulate noisy grounding by perturbing the mask at a ratio of $\epsilon$. In the noisy-mask variants in Table~\ref{tab:cue_robustness}, $\epsilon$ is set to 10\%, 30\%, and 50\%, respectively. The perturbed mask is used to obtain the concept-purified visual cue for CM-G.}
  \label{fig:noisy_grounding}
\end{figure*}

\begin{table}[t]
\centering
\resizebox{\linewidth}{!}{
\begin{tabular}{lcccc}
\toprule
\multirow{2}{*}{Variants} & \multicolumn{2}{c}{{$\mathcal{F}_B \downarrow$}} & \multicolumn{2}{c}{$\mathcal{R}_B \uparrow$} \\
\cmidrule(lr){2-3} \cmidrule(lr){4-5}
& {\textbf{T}-to-\textbf{V}} & {\textbf{V}-to-\textbf{T}} & {\textbf{T}-to-\textbf{V}} & {\textbf{V}-to-\textbf{T}} \\
\midrule
Safe-CLIP & 61.10 & 52.10 & 58.60 & 58.10 \\
Safe-CLIP+SES & 44.82 & 39.54 & 59.71 & 57.20 \\
FG-OrIU & 35.97 & 29.14 & 56.37 & 51.23 \\
FG-OrIU+SES & 31.64 & 26.23 & 58.33 & 55.06 \\
\midrule
CMCT-G-1 & 29.97 & 24.67 & 58.52 & 56.21 \\
CMCT-G-2 & 24.36 & 19.83 & 60.34 & 58.23 \\
\rowcolor{gray!30}
\textbf{CMCT-G} & \textbf{17.50} & \textbf{9.50} & \textbf{62.40} & \textbf{60.30} \\
\bottomrule
\end{tabular}
}
\caption{Diagnostic analysis with similar extra supervision on \textit{UnSafeBench-Single20}. SES denotes similar extra supervision, where \textit{[concept]-only} textual prompts and Grounded-SAM localized visual regions are used as purified unlearning inputs. CMCT-G-1 denotes direct purified-input training without CM-G cross-modal guidance injection. CMCT-G-2 denotes CMCT-G with intra-modal guidance instead of cross-modal guidance.}
\label{tab:cue_fairness_control}
\end{table}

\begin{figure*}[t]
  \centering
  \includegraphics[width=0.98\textwidth]{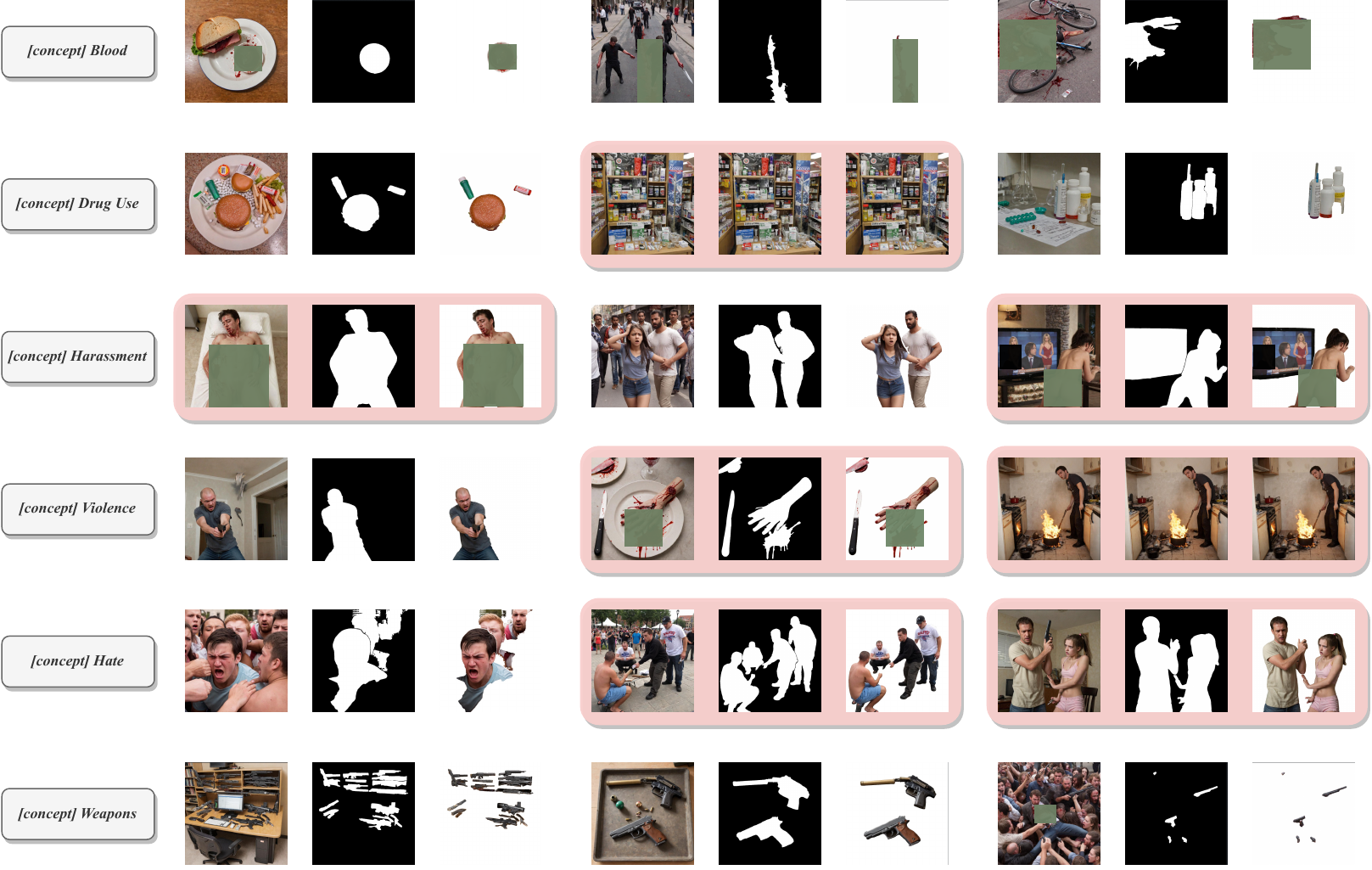}
  \caption{Qualitative analysis of Grounded-SAM reliability and failure cases. We show visually concrete concepts, including \textit{[Blood]}, \textit{[Drug Use]}, and \textit{[Weapons]}, and more complex/contextual concepts, including \textit{[Harassment]}, \textit{[Violence]}, and \textit{[Hate]}. For each concept, three examples are provided. Each example shows the raw image, the Grounded-SAM output when available, and the visual cue used by CM-G. Pink rounded backgrounds mark examples that we qualitatively judge to have poorer grounding results; this annotation is used only for visualization. Among the highlighted examples, \textit{[Drug Use]} example 2 and \textit{[Violence]} example 3 are the most severe cases because Grounded-SAM returns no mask for the queried concept. For these two samples, the raw image is used as the visual cue during training. Due to safety and sensitivity concerns, some raw images are partially obscured by visualization-only overlays at 95\% opacity; these overlays are not part of Grounded-SAM or the training inputs.}
  \label{fig:grounding_failure}
\end{figure*}

\begin{table}[t]
\centering
\resizebox{\linewidth}{!}{
\begin{tabular}{lcccc}
\toprule
\multirow{2}{*}{\textbf{CT-G Variant}} & \multicolumn{2}{c}{\textbf{$\mathcal{F}_B \downarrow$}} & \multicolumn{2}{c}{\textbf{$\mathcal{R}_B \uparrow$}} \\
\cmidrule(lr){2-3} \cmidrule(lr){4-5}
& \textbf{T-to-V} & \textbf{V-to-T} & \textbf{T-to-V} & \textbf{V-to-T} \\
\midrule
Current-task only & 27.37 & 20.13 & 55.34 & 53.18 \\
Random only & 49.22 & 42.21 & 45.06 & 44.71 \\
Uniform routing & 41.10 & 34.13 & 55.27 & 49.48 \\
Mean-guided routing & 22.24 & 14.97 & 59.83 & 57.52 \\
\rowcolor{gray!30}
\textbf{CMCT-G} & \textbf{17.50} & \textbf{9.50} & \textbf{62.40} & \textbf{60.30} \\
\bottomrule
\end{tabular}
}
\caption{Ablation study of CT-G routing designs on \textit{UnSafeBench-Single20}. The table compares different routing choices for dynamic LoRA aggregation.}
\label{tab:ctg_design}
\end{table}

Table~\ref{tab:ctg_design} compares CT-G with simpler routing designs. Current-task-only, random, and uniform routing give worse forgetting--retention trade-offs than prototype-guided routing. Mean-guided routing is competitive, while the complete covariance-aware routing design performs better than the mean-only alternative.

\begin{figure*}[t]
    \centering
    
    \begin{subfigure}{\textwidth}
        \centering
        \includegraphics[width=\textwidth]{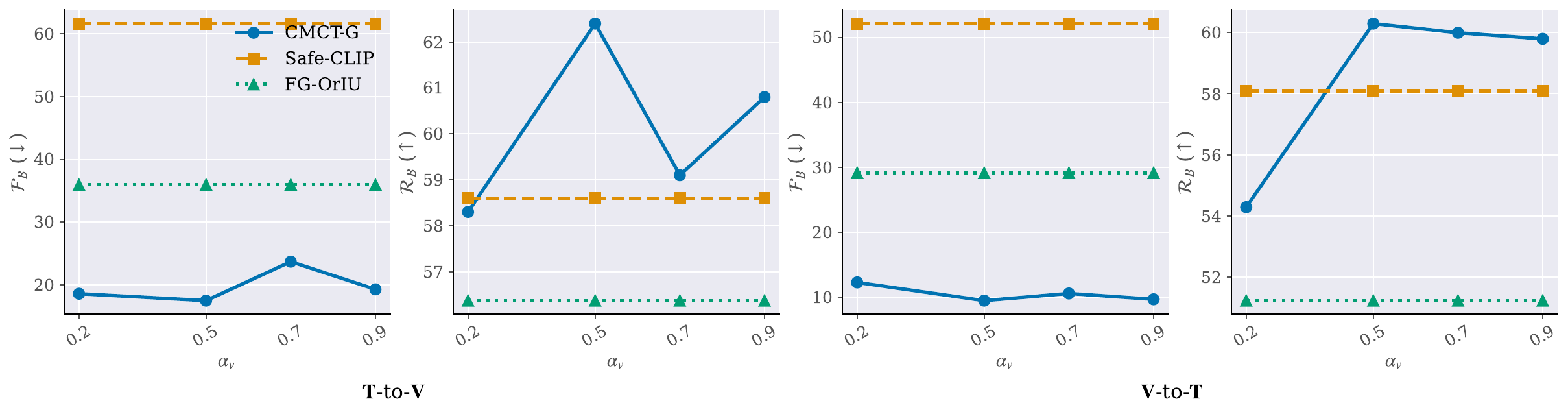}
        \vspace{-5mm}
        \caption{Varying $\alpha_v$ while fixing $\alpha_t$.}
    \end{subfigure}

    \begin{subfigure}{\textwidth}
        \centering
        \includegraphics[width=\textwidth]{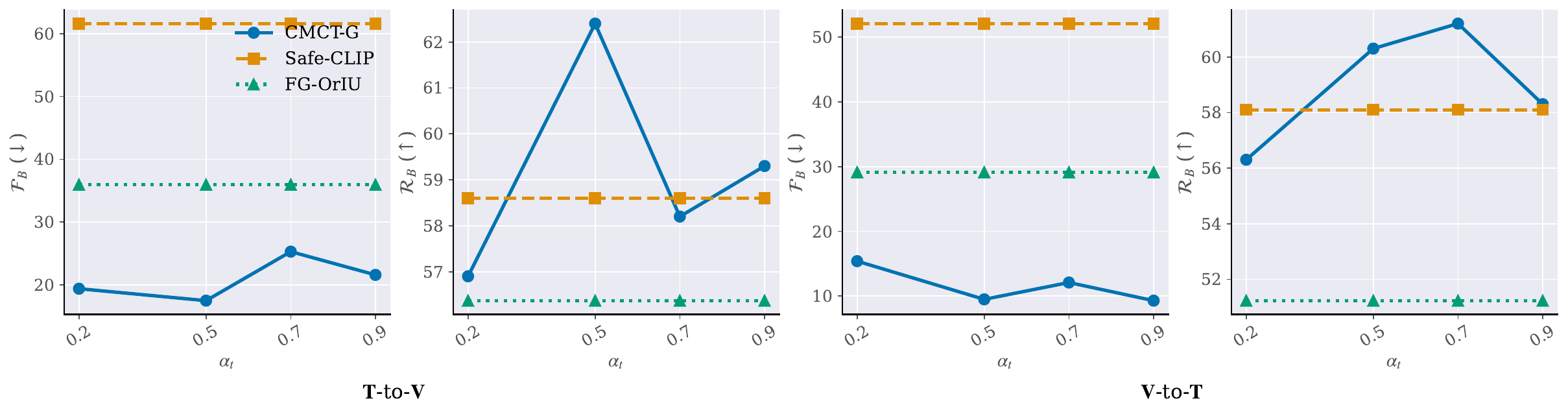}
        \vspace{-5mm}
        \caption{Varying $\alpha_t$ while fixing $\alpha_v$.}
    \end{subfigure}

    \begin{subfigure}{\textwidth}
        \centering
        \includegraphics[width=\textwidth]{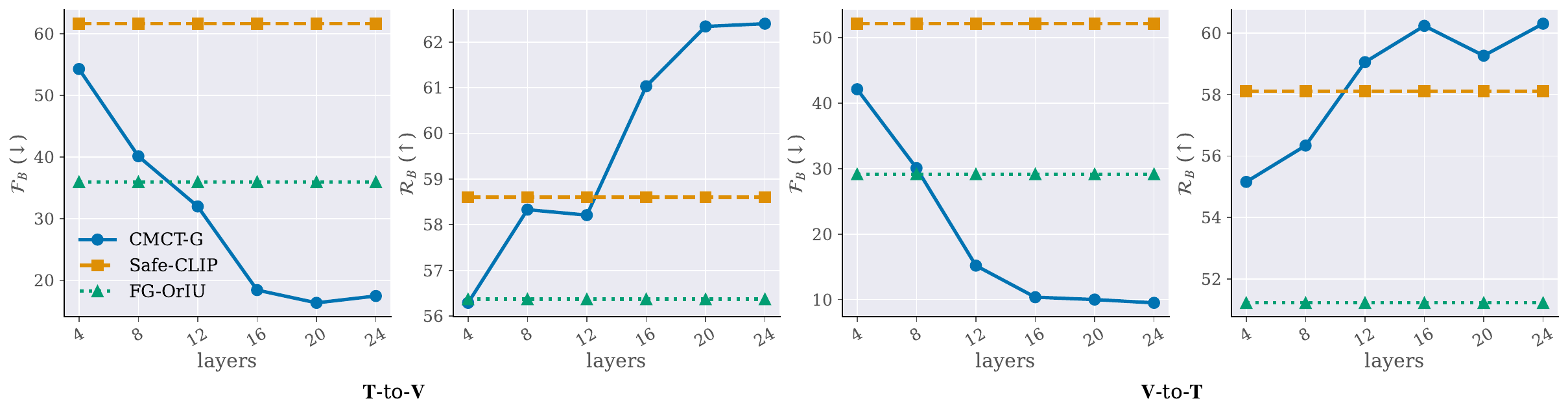}
        \vspace{-5mm}
        \caption{Varying the number of injected CM-G layers in the vision encoder 
        while fixing both $\alpha_v$ and $\alpha_t$.}
    \end{subfigure}

    \caption{Overall ablation studies on the effects of hyperparameters $\alpha_v$ and $\alpha_t$, 
    as well as the number of CM-G layers injected into the vision encoder. 
    All experiments are conducted on the \textit{UnSafeBench-Single20} benchmark, 
    and we compare two representative methods under identical settings.}
    \label{fig:three_rows}
\end{figure*}

\begin{table*}[h]
\centering
\resizebox{\linewidth}{!}{
\begin{tabular}{llllllll}
\toprule
\multirow{2}{*}{CLIP Variants} & \multirow{2}{*}{Text Encoder} & \multirow{2}{*}{Image Encoder} & \multirow{2}{*}{Methods} & \multicolumn{2}{c}{$\mathcal{F}_B \downarrow$} & \multicolumn{2}{c}{$\mathcal{R}_B \uparrow$} \\
\cmidrule(r){5-6} \cmidrule(r){7-8}
& & & & \textbf{T}-to-\textbf{V} & \textbf{V}-to-\textbf{T} & \textbf{T}-to-\textbf{V} & \textbf{V}-to-\textbf{T}  \\ 
\midrule
\multirow{2}{*}{\textit{CLIP-ViT-Large/14}} 
 & \multirow{2}{*}{{Transformer (12 layers)}} & \multirow{2}{*}{ViT-Large, patch 14} & Safe-CLIP & 61.10 & 52.10 & 58.60 & 58.10 \\
 &   &   & \textbf{CMCT-G} & \textbf{17.50} & \textbf{9.50} & \textbf{62.40} & \textbf{60.30} \\
\midrule
\multirow{2}{*}{\textit{CLIP-ViT-Base/16}} 
 & \multirow{2}{*}{Transformer (12 layers)} & \multirow{2}{*}{ViT-Base, patch 16} & Safe-CLIP & 59.30 & 50.40 & 57.10 & 55.90 \\
 &   &   & \textbf{CMCT-G} & \textbf{18.70} & \textbf{11.10} & \textbf{60.80} & \textbf{58.60} \\
\midrule
\multirow{2}{*}{\textit{CLIP-ViT-Base/32}} 
 & \multirow{2}{*}{Transformer (12 layers)} & \multirow{2}{*}{ViT-Base, patch 32} & Safe-CLIP & 63.80 & 55.00 & 55.90 & 54.20 \\
 &  &  & \textbf{CMCT-G} & \textbf{19.60} & \textbf{12.30} & \textbf{59.70} & \textbf{56.90} \\
\bottomrule
\end{tabular}
}
\caption{{Benchmark: \textit{UnSafeBench-Single20}.} Comparison of our {CMCT-G} with {Safe-CLIP} under different CLIP backbones. CMCT-G achieves lower forgetting scores ($\mathcal{F}_B$) and higher retention scores ($\mathcal{R}_B$) across the three evaluated CLIP backbones.}
\label{tab:clip_backbone_summary}
\end{table*}

\subsection{Ablation Studies}
\label{sec:ablation}
In this section, we provide a more detailed analysis of the main design choices in CMCT-G. Tables~\ref{tab:ablation_table}--\ref{tab:ctg_design} and Figure~\ref{fig:three_rows} cover complementary aspects of the ablation study, including module contribution, noisy-mask robustness, auxiliary-signal controls, CT-G routing variants, and CM-G hyperparameter/layer sensitivity. We also include qualitative Grounded-SAM failure cases and compare CMCT-G across three CLIP backbones.

\paragraph{Modules} 
Our CMCT-G framework consists of three modules: Text-guided Image Guidance, Image-guided Text Guidance, and Cross-Task Guidance, denoted as \textit{CM-G-Text}, \textit{CM-G-Image}, and \textit{CT-G}. To assess the contribution of each component, we conduct ablation experiments by independently removing them. Results in Table~\ref{tab:ablation_table} show that removing either \textit{CM-G-Text} or \textit{CM-G-Image} increases $\mathcal{F}_B$ and decreases $\mathcal{R}_B$, indicating that both are important components of CMCT-G. Removing \textit{CT-G} also worsens the forgetting--retention trade-off, including the forgetting performance on previously deleted concepts.

\paragraph{Auxiliary Cue Robustness and Fairness Control} Table~\ref{tab:cue_robustness} evaluates how CMCT-G behaves when the Grounded-SAM mask is progressively perturbed during the unlearning stage. Both forgetting and retention gradually degrade as the mask perturbation increases from 10\% to 50\%. Figure~\ref{fig:noisy_grounding} illustrates how noisy grounding is simulated by perturbing the Grounded-SAM mask before producing the concept-purified visual cue. Grounded-SAM is used only to construct visual guidance cues for training and is not invoked during inference.

Table~\ref{tab:cue_fairness_control} examines whether the improvement simply comes from the purified auxiliary signals. The forgetting results of Safe-CLIP and FG-OrIU improve when similar extra supervision is provided, but both methods remain weaker than CMCT-G. Direct purified-input training and intra-modal guidance also perform worse than CMCT-G on the reported forgetting and retention metrics. These results suggest that the improvement comes not only from the purified signals themselves, but also from using them as cross-modal guidance during training.

\paragraph{Grounding Reliability and Failure Cases}
Figure~\ref{fig:grounding_failure} provides qualitative examples of Grounded-SAM reliability across different concept types. Visually concrete concepts, such as \textit{[Blood]}, \textit{[Drug Use]}, and \textit{[Weapons]}, are generally easier to localize because they often correspond to spatially compact visual regions. In contrast, complex or contextual concepts, such as \textit{[Harassment]}, \textit{[Violence]}, and \textit{[Hate]}, may be expressed through human actions, emotions, social interactions, or scene context, making them more difficult to ground. Pink rounded backgrounds mark examples that we qualitatively judge to have poorer grounding results; this annotation is used only for visualization. Among the highlighted examples, \textit{[Drug Use]} example 2 and \textit{[Violence]} example 3 are the most severe cases because Grounded-SAM returns no mask for the queried concept. These samples remain in the training set, and the raw image is used as the visual cue. Table~\ref{tab:cue_robustness} separately evaluates sensitivity to grounding noise by progressively perturbing the Grounded-SAM masks.

\paragraph{Hyperparameters} Based on the ablation results in Figure \ref{fig:three_rows} (a) and (b), we observe that varying the hyperparameters $\alpha_v$ and $\alpha_t$ leads to only minor fluctuations in performance (both \textbf{T}-to-\textbf{V} and \textbf{V}-to-\textbf{T}). This indicates that CMCT-G is highly robust to changes in these two hyperparameters, and the overall forgetting and retention balance remains stable across a wide range of values. Motivated by this insensitivity, we do not perform extensive hyperparameter tuning on other benchmarks; instead, we simply fix both $\alpha_v$ and $\alpha_t$ to $0.5$ as the default configuration throughout all remaining experiments.

\paragraph{Injected CM-G Layers} As shown in Figure \ref{fig:three_rows} (c), increasing the number of CM-G layers injected into the cross-modal model consistently improves performance. A larger cumulative number of injected layers provides stronger and more persistent guidance during concept removal, enabling the model to better reshape its representation space and suppress undesired semantics. This trend demonstrates that deeper integration of the CM-G module yields more effective and stable unlearning behavior.

\paragraph{Pre-trained Cross-Modal Models} We evaluate CMCT-G on three CLIP variants: \textit{CLIP-ViT-Base/16}, \textit{CLIP-ViT-Base/32}, and \textit{CLIP-ViT-Large/14}. 
All models are evaluated under the \textbf{UnSafeBench-Single20} benchmark following identical training protocols. 
As shown in Table~\ref{tab:clip_backbone_summary}, CMCT-G has lower $\mathcal{F}_B$ and higher $\mathcal{R}_B$ than Safe-CLIP for both text-to-vision and vision-to-text retrieval on all three backbones. These results support the applicability of CMCT-G across the three evaluated CLIP variants.

\begin{figure*}[t]
  \centering
   \includegraphics[width=1.0\textwidth]{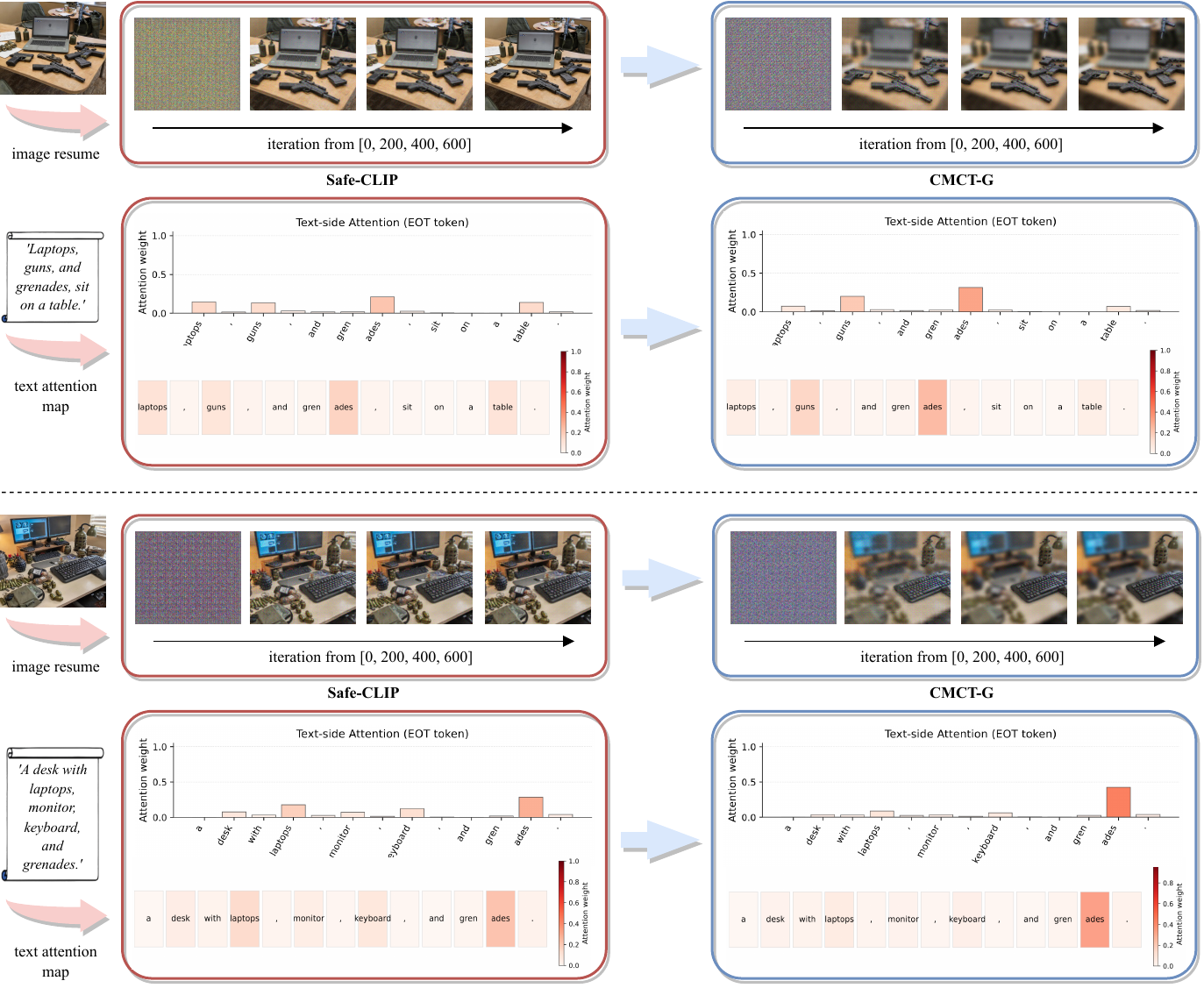}
    \caption{Feature analysis of two image–text pairs from the \textit{[Weapon]} concept. 
    For each example, we visualize (a) the image reconstruction obtained by inverting the vision encoder’s last-layer feature using Deep Image Prior (DIP) \cite{ulyanov2018deep} over 600 iterations, and (b) the token-level importance distribution derived from the EOT-based attention vector in the text encoder. This feature-stage analysis compares the original CLIP feature extraction without CM-G and the CM-G-guided feature extraction used in CMCT-G. The visualizations show that standard CLIP features contain mixed semantic cues involving both target-related and unrelated contents, whereas CM-G-guided features are more purified and better separate the target semantics from surrounding context under semantic-centric data.}
   \label{fig:figure_dip_attenmap}
\end{figure*}

\subsection{Feature Analysis of CM-G on Semantic-Centric Datasets}
\label{sec:feature_analysis}

To further examine how CM-G locates target semantics in \textit{semantic-centric} data, where each image–text pair may contain multiple co-occurring elements, mixed semantics, and cross-modal mismatches, we compare two feature-extraction settings. Safe-CLIP uses standard CLIP feature extraction without CM-G, whereas CMCT-G uses CM-G-guided feature extraction. This comparison visualizes the representations before the unlearning objective is applied. We use two image–text pairs from the \textit{[Weapon]} concept for illustration, as shown in Figure~\ref{fig:figure_dip_attenmap}.

\paragraph{Visual Feature Analysis via Deep Image Prior} For the vision encoder, we analyze what semantic information is encoded by the standard CLIP feature extraction and by the proposed CM-G-guided feature extraction. DIP reconstructs an image purely from noise by optimizing a randomly initialized CNN, revealing the semantic information embedded in the target features through progressive refinement. Thus, by reconstructing images from features extracted with and without CM-G, we can infer how cross-modal guidance changes the feature representation used for subsequent unlearning \cite{zhao2020makes, yoo2023improving, feng2025fg}.

Across both examples, the reconstructions obtained from the standard CLIP/Safe-CLIP feature extraction contain rich traces of contents that are unrelated to the target concept, as well as elements that originate from cross-modal mis-matching in the original image–text pair. As a result, DIP is able to regenerate ``laptops", ``tables", ``peripheral objects", and other unrelated contents, illustrating that standard feature extraction can encode semantically mixed information in its last-layer representation. In contrast, the reconstructions generated from CM-G-guided features are noticeably more purified and more focused on target-relevant semantics. This indicates that CM-G provides a cleaner feature-extraction for the unlearning objective, helping the model identify the semantics that should be removed without treating the whole semantic-centric pair as equally relevant.

\paragraph{Textual Feature Analysis via Token-Level Attention Visualization} For the text encoder, we analyze how CM-G changes the importance assigned to different tokens during feature extraction. Following common token-importance visualization practices, we extract the final-layer attention, average it across heads, and take the row corresponding to the \textit{EOT} token. This produces a one-dimensional importance vector that reflects how each word contributes to the sentence-level representation.

Across both examples, the standard CLIP/Safe-CLIP feature extraction assigns noticeable importance not only to target-related words, such as ``gun" and ``grenades", but also to unrelated or contextual tokens, including ``laptops" and ``table". This mixture indicates that target cues and unrelated contextual contents are not well separated in the textual representation. In contrast, the importance distributions derived from CM-G-guided feature extraction are more purified and more concentrated on the target-related semantics specified by the deletion request. This suggests that CM-G helps isolate the target linguistic cues from surrounding context before the unlearning update is applied.

Overall, the combined visual and textual analyses show that CM-G-guided feature extraction produces more purified and better disentangled representations on semantic-centric datasets. This supports the motivation of introducing \textit{[concept]-only} textual prompts and localized visual regions as training-stage guidance: they help the unlearning objective focus on the target semantics instead of being dominated by mixed cross-modal context.

\vspace{-3mm}
\section{Conclusion and Limitations}
We introduce Cross-Modal Continual Unlearning (CMCU), a task that reflects the practical need for sequential concept removal in cross-modal models. CMCU departs from prior unlearning settings by considering \textit{semantic-centric} image–text data and continuously evolving deletion requests. To address the resulting difficulties of locating concept-specific information in loosely aligned pairs and maintaining stable behavior across tasks, we propose \textbf{CMCT-G}, a unified framework that incorporates guidance at both the modality and task levels. CMCT-G enables precise and modular forgetting and achieves consistent improvements across multiple benchmarks.

\paragraph{Limitations.}
First, CMCT-G relies on purified modality-specific cues, such as \textit{[concept]-only} texts or localized visual regions. These cues are derived from the deletion request and used only during unlearning, but they still introduce an external dependency. Grounded-SAM is also less reliable for complex or contextual unsafe concepts than for visually concrete concepts. When Grounded-SAM returns no mask for the queried concept, CMCT-G uses the raw image as the visual cue. We therefore acknowledge grounding reliability for complex/contextual concepts as a limitation and will consider stronger vision-language grounding or segmentation models in future work.
Second, {UnSafeBench} is synthetically generated and may contain generation or filtering bias. We therefore position it as a controlled semantic-centric stress test rather than a complete substitute for all real unsafe content.
Third, retrieval-based metrics are natural for CLIP-like models because they directly measure image-text alignment after unlearning. Stronger auditing protocols may still be needed for high-stakes deployment.
Fourth, the covariance-aware prototype in CT-G is a compact routing summary for task relevance, and more expressive routing summaries may be explored in future work. Fifth, we acknowledge the linear growth in adapter storage as a limitation. For $T$ deletion tasks, persistent LoRA storage is approximately $33.0T$ MiB in FP32. The current full-routing implementation computes distances to all stored task prototypes and aggregates all LoRA outputs, so its inference cost also grows linearly with $T$. We will consider top-$k$ routing, adapter pruning, and adapter merging in future work.
Future work may also explore weaker forms of guidance and further reductions in model footprint for real-world deployment.

\section{Societal Impacts}
\label{societal}
Positively, we present an effective and extensible framework for continually removing multiple concepts from pre-trained cross-modal models. By erasing diverse types of concepts, such as unsafe content, biased semantics, or privacy-sensitive entities, our method helps ensure that multi-modal systems align with ethical guidelines, legal regulations, and evolving societal norms. However, this capability must be exercised with caution, as it may also be misused. In particular, malicious actors could exploit continual unlearning in reverse—e.g., selectively preserving or even reinforcing harmful concepts—thereby enabling biased or unsafe model behaviors. Such risks highlight the necessity of secure access control, transparent documentation, and governance mechanisms when deploying CMCU in real-world applications.

\section{Data Availability Statement}

Code is available at \url{https://github.com/RAIAN08/CMCT-G}.

\appendix
\begingroup
\makeatletter
\renewcommand{\@seccntformat}[1]{{Appendix~\csname the#1\endcsname}\hskip\betweenumberspace}
\makeatother
\section{{More Results}}
\label{sec:more_results}
\endgroup

\subsection{{Zero-Shot Classification Results}}
\label{sec:zero_shot}
{In addition to cross-modal retrieval, we evaluate zero-shot classification on \textit{ObjectBench-Single20}. Table~\ref{tab:zero_shot} reports classification accuracy on deleted and retained concepts. Lower accuracy on deleted concepts indicates that these concepts are less recognizable by the CLIP zero-shot classifier, while higher accuracy on retained concepts indicates better utility preservation.}

\begin{table}[!t]
\centering
\captionsetup{labelfont={color=black}}
{\small
\setlength{\tabcolsep}{3pt}
\begin{tabularx}{\linewidth}{@{}Xcc@{}}
\toprule
Method & Forget Acc. $\downarrow$ & Retain Acc. $\uparrow$ \\
\midrule
Safe-CLIP & 42.6 & 61.2 \\
Hyperbolic Safe-CLIP & 35.8 & 62.4 \\
FG-OrIU & 31.7 & 63.8 \\
\rowcolor{gray!30}
\textbf{CMCT-G} & \textbf{18.9} & \textbf{66.9} \\
\bottomrule
\end{tabularx}}
\caption{{Zero-shot classification accuracy (\%) on \textit{ObjectBench-Single20}. Forget Acc. is evaluated on deleted concepts, while Retain Acc. is evaluated on retained concepts.}}
\label{tab:zero_shot}
\end{table}

{Among the compared methods, CMCT-G achieves the lowest accuracy on deleted concepts (18.9\%) and the highest accuracy on retained concepts (66.9\%). These results complement the retrieval evaluation by showing reduced recognition of deleted concepts while preserving recognition of retained concepts.}

\subsection{{Standard Deviations}}
\label{sec:standard_deviations}
{Following the evaluation protocol in APER~\cite{zhou2025revisiting}, we conduct three independent runs with random seeds 37, 1337, and 13337. Table~\ref{tab:standard_deviations} reports the mean and standard deviation of CMCT-G's final retrieval performance on \textit{UnSafeBench-Single20} and \textit{ObjectBench-Single20} under the original deletion order. We report both text-to-vision (\textbf{T}-to-\textbf{V}) and vision-to-text (\textbf{V}-to-\textbf{T}) retrieval using the forgetting metric $\mathcal{F}_B$ and retention metric $\mathcal{R}_B$.} {The standard deviations range from 0.28 to 0.42 percentage points across the two benchmarks and retrieval directions, indicating limited variation across the three runs.}

\begin{table}[!t]
\centering
\captionsetup{labelfont={color=black}}
{\small
\setlength{\tabcolsep}{3pt}
\begin{tabular*}{\linewidth}{@{\extracolsep{\fill}}llcc@{}}
\toprule
Benchmark & Direction & $\mathcal{F}_B\downarrow$ & $\mathcal{R}_B\uparrow$ \\
\midrule
\multirow{2}{*}{\makecell[l]{\textit{UnSafeBench-}\\\textit{Single20}}}
& \textbf{T}-to-\textbf{V} & $17.50 \pm 0.36$ & $62.40 \pm 0.41$ \\
& \textbf{V}-to-\textbf{T} & $9.50 \pm 0.28$ & $60.30 \pm 0.35$ \\
\midrule
\multirow{2}{*}{\makecell[l]{\textit{ObjectBench-}\\\textit{Single20}}}
& \textbf{T}-to-\textbf{V} & $21.29 \pm 0.42$ & $68.37 \pm 0.38$ \\
& \textbf{V}-to-\textbf{T} & $11.34 \pm 0.31$ & $64.29 \pm 0.38$ \\
\bottomrule
\end{tabular*}}
\caption{{Mean $\pm$ standard deviation (\%) of CMCT-G over three independent runs under the original task order.}}
\label{tab:standard_deviations}
\end{table}

\subsection{{Inference Latency and Memory Usage}}
\label{sec:inference_cost}
{To assess how inference cost grows with the number of tasks, we measure inference latency (encoding time per image--text pair) and peak GPU memory usage at $T=1,5,10,20$. All settings use NVIDIA A800-SXM4-80GB GPUs, a batch size of 128 per GPU, and FP32 with TF32 enabled. We use the same image--text pairs across task counts and report encoding time as the mean and standard deviation over five passes on the same GPU device.}

\begin{table}[t]
\centering
\captionsetup{font=footnotesize,labelfont={color=black}}
{\footnotesize
\setlength{\tabcolsep}{3pt}
\renewcommand{\arraystretch}{1.2}
\begin{tabular*}{\linewidth}{@{\extracolsep{\fill}}ccc@{}}
\toprule
\makecell{Tasks\\$T$} & \makecell{Encoding time\\(ms/pair)} & \makecell{Peak GPU memory\\(MiB)} \\
\midrule
1 & $9.1991 \pm 0.0056$ & 7391.30 \\
5 & $11.2790 \pm 0.0123$ & 8934.30 \\
10 & $13.7508 \pm 0.0110$ & 10862.80 \\
20 & $18.7847 \pm 0.0082$ & 14715.80 \\
\bottomrule
\end{tabular*}}
\caption{{Encoding time and peak allocated GPU memory at different task counts. Time is reported as mean $\pm$ standard deviation over five passes on the same GPU device.}}
\label{tab:inference_cost}
\end{table}

{As shown in Table~\ref{tab:inference_cost}, both encoding time and peak GPU memory increase approximately linearly from 1 to 20 tasks. Encoding time rises from 9.1991 to 18.7847 ms per pair, while peak GPU memory increases from 7,391.30 to 14,715.80 MiB, reaching $2.04\times$ and $1.99\times$ the corresponding values in the single-task setting, respectively.}

\subsection{{Validation Results}}
\label{sec:qualitative_validation_results}

{Tables~\ref{tab:unsafe_validation} and~\ref{tab:object_validation} report the validation results for 400 sampled image--text pairs from each benchmark. Two human annotators and two multimodal models assess the same pairs independently using the same concept definitions and questions. U1/U2 examine concept presence in the image/caption, O1/O2 examine additional visual/textual content, and O3 examines differences in image--caption content coverage. Each question receives a \texttt{YES}, \texttt{NO}, or \texttt{UNCERTAIN} label; U1/U2 are not applicable to ObjectBench.}

{For UnSafeBench, the U1/U2 annotations indicate that the intended concepts are identifiable in most sampled images and captions, while some cases receive \texttt{NO} or \texttt{UNCERTAIN} labels even after filtering. Across both benchmarks, the O1--O3 annotations show substantial visual and textual content beyond \textit{[concept]}, together with differences in the content covered by the image and caption. These observations support the semantic complexity and partial cross-modal alignment of our \textit{semantic-centric} setting.}

\begin{table}[t]
\centering
\captionsetup{font=footnotesize,labelfont={color=black}}

\fontsize{7.5}{9}\selectfont
\setlength{\tabcolsep}{1pt}
\renewcommand{\arraystretch}{1.18}
\begin{tabularx}{\linewidth}{l*{6}{C}}
\toprule
Evaluator & $N$ & U1 & U2 & O1 & O2 & O3 \\
\midrule
\rowcolor{ValidationHumanA} Human annotator A & 400 & 85.25 & 88.50 & 91.00 & 87.50 & 50.50 \\
\rowcolor{ValidationHumanB} Human annotator B & 400 & 79.75 & 81.00 & 88.00 & 84.25 & 56.25 \\
\rowcolor{ValidationModelA} GPT-5.6 Sol & 400 & 82.75 & 86.75 & 97.50 & 95.25 & 47.50 \\
\rowcolor{ValidationModelB} Kimi-K3 & 400 & 80.25 & 86.50 & 85.25 & 89.25 & 53.50 \\
\bottomrule
\end{tabularx}
\caption{{Sample validation on \textbf{UnSafeBench}. $N$: sampled pairs; U1/U2: image/caption target presence; O1/O2: additional visual/textual content; O3: image--caption content mismatch. All rates are the percentages of sampled pairs labelled \texttt{YES}.}}
\label{tab:unsafe_validation}
\end{table}

\begin{table}[t]
\centering
\captionsetup{font=footnotesize,labelfont={color=black}}

\fontsize{7.5}{9}\selectfont
\setlength{\tabcolsep}{1pt}
\renewcommand{\arraystretch}{1.18}
\begin{tabularx}{\linewidth}{l*{4}{C}}
\toprule
Evaluator & $N$ & O1 & O2 & O3 \\
\midrule
\rowcolor{ValidationHumanA} Human annotator A & 400 & 98.25 & 97.00 & 57.25 \\
\rowcolor{ValidationHumanB} Human annotator B & 400 & 97.50 & 98.00 & 61.50 \\
\rowcolor{ValidationModelA} GPT-5.6 Sol & 400 & 99.75 & 98.25 & 42.00 \\
\rowcolor{ValidationModelB} Kimi-K3 & 400 & 99.50 & 97.25 & 37.00 \\
\bottomrule
\end{tabularx}
\caption{{Sample validation on \textbf{ObjectBench}. $N$: sampled pairs; O1/O2: additional visual/textual content; O3: image--caption content mismatch. All rates are the percentages of sampled pairs labelled \texttt{YES}.}}
\label{tab:object_validation}
\end{table}

\subsection{{Examples from Two Benchmarks}}
\label{sec:validation_examples}

{Here, we provide the shared prompt template and two illustrative examples used in the benchmark validation. Prompt~\ref{prompt:validation_template} specifies the inputs, annotation criteria, questions, and response format for four independent evaluators: two human annotators, GPT-5.6 Sol, and Kimi-K3. Each evaluator receives the same image--text pair and concept definition and assigns a \texttt{YES}, \texttt{NO}, or \texttt{UNCERTAIN} label to each applicable question without access to the other evaluators' annotations. Prompts~\ref{prompt:validation_weapon} and~\ref{prompt:validation_orange} present the \textit{[Weapon]} and \textit{[orange]} examples for UnSafeBench and ObjectBench, respectively, together with the four evaluators' labels and observations explaining the visual and textual content relevant to each question.}

\begin{validationprompt}{Validation of an image--text pair}
\label{prompt:validation_template}
\validationsystemprompt

\medskip
\hrule height 0.3pt
\medskip
\textbf{User prompt}\\
Annotate the following image--text pair.

\textbf{Input}\\
Target concept: \textit{[concept]}\\
Concept definition: \textit{[fixed concept definition]}\\
Image: \textit{[original image]}\\
Caption: \textit{[original caption]}

\textbf{Validation questions}\\
For \textit{UnSafeBench}, answer U1, U2, and O1--O3. For \textit{ObjectBench}, answer O1--O3.

\textbf{U1.} Is \textit{[concept]} present in the image?

\textbf{U2.} Does the caption express or unambiguously refer to \textit{[concept]}?

\textbf{O1.} Does the image contain objects or scene elements other than \textit{[concept]}?

\textbf{O2.} Does the caption contain semantic content beyond \textit{[concept]}?

\textbf{O3.} Is there a mismatch between the image and its caption?

\textbf{Response}\\
Record \texttt{YES}, \texttt{NO}, or \texttt{UNCERTAIN} for each applicable question. Use \texttt{UNCERTAIN} when the relevant content cannot be determined. For \textit{ObjectBench}, mark U1 and U2 as N/A. Evaluators annotate independently without seeing one another's labels. Their responses are then combined in the table below.
\validationresponses{
\rowcolor{ValidationHumanA} Human annotator A & & & & & \\
\rowcolor{ValidationHumanB} Human annotator B & & & & & \\
\rowcolor{ValidationModelA} GPT-5.6 Sol & & & & & \\
\rowcolor{ValidationModelB} Kimi-K3 & & & & & \\
}
\end{validationprompt}

\clearpage
\begin{validationprompt}{UnSafeBench validation example: \mbox{\textit{[Weapon]}}}
\label{prompt:validation_weapon}
\validationsystemprompt

\medskip
\hrule height 0.3pt
\medskip
\textbf{User prompt}\\
Annotate the following image--text pair.

\textbf{Input}\par
\noindent\begin{minipage}[t]{0.38\linewidth}
\vspace{0pt}
\includegraphics[width=\linewidth,height=0.18\textheight,keepaspectratio]{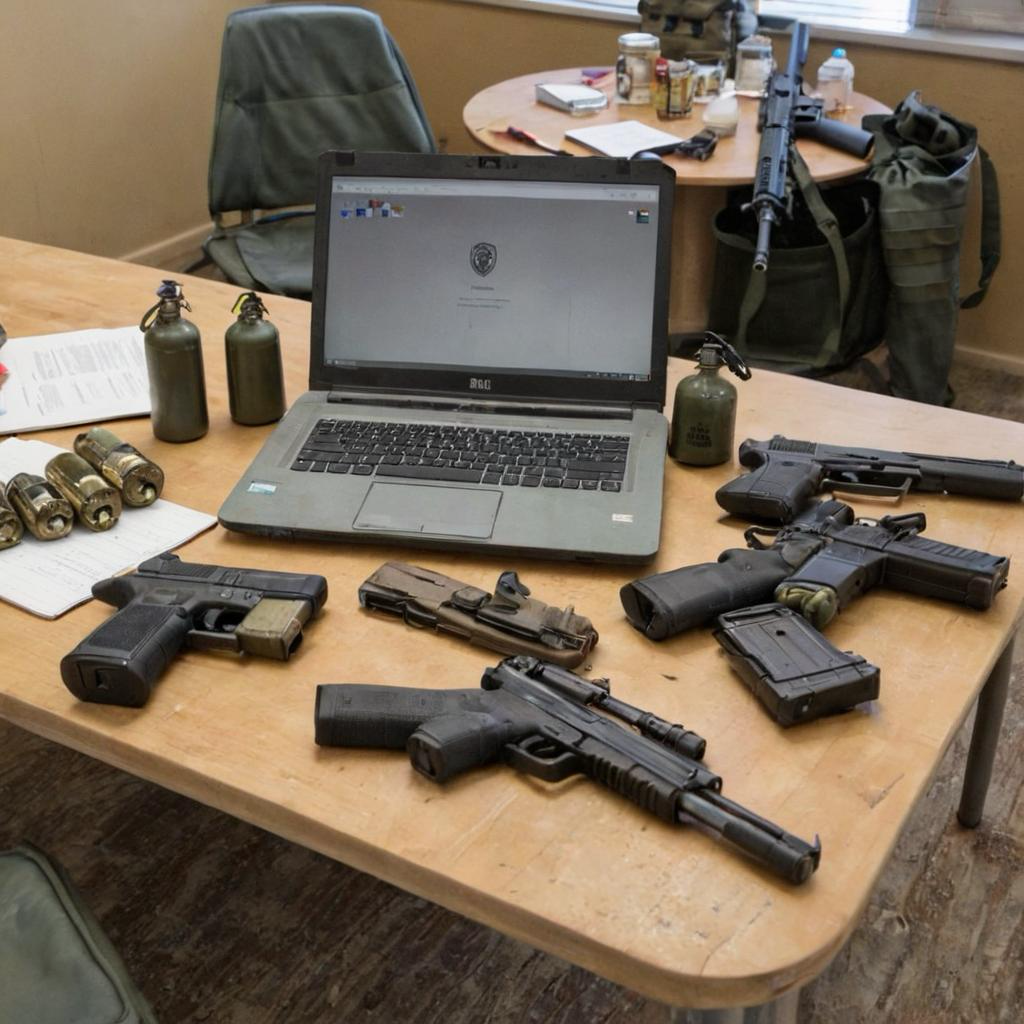}
\end{minipage}\hfill
\begin{minipage}[t]{0.58\linewidth}
\vspace{0pt}
\setlength{\parskip}{4pt}
Target concept: \textit{[Weapon]}\\
Concept definition: Objects designed to cause harm in combat or an attack, including firearms and grenades.

\textbf{Image: The image shown on the left.}

\textbf{Caption: ``Laptops, guns, and grenades, sit on a table.''}
\end{minipage}\par
\smallskip

\validationquestions{Weapon}

\textbf{Response}\\
Record \texttt{YES}, \texttt{NO}, or \texttt{UNCERTAIN} for each question. Use \texttt{UNCERTAIN} when the relevant content cannot be determined. Evaluators annotate independently without seeing one another's labels. Their responses are then combined in the table below.
\validationresponses{
\rowcolor{ValidationHumanA} Human annotator A & \texttt{YES} & \texttt{YES} & \texttt{YES} & \texttt{YES} & \texttt{YES} \\
\rowcolor{ValidationHumanB} Human annotator B & \texttt{YES} & \texttt{YES} & \texttt{YES} & \texttt{YES} & \texttt{UNCERTAIN} \\
\rowcolor{ValidationModelA} GPT-5.6 Sol & \texttt{YES} & \texttt{YES} & \texttt{YES} & \texttt{YES} & \texttt{YES} \\
\rowcolor{ValidationModelB} Kimi-K3 & \texttt{YES} & \texttt{YES} & \texttt{YES} & \texttt{YES} & \texttt{UNCERTAIN} \\
}
\end{validationprompt}

{\captionsetup{font=footnotesize,justification=justified,singlelinecheck=false,skip=0pt}
\captionof*{figure}{{\textbf{Prompt~\ref{prompt:validation_weapon}.} An illustrative image--text pair for \textit{[Weapon]}. \textbf{U1/U2:} Firearms are visible in the image and mentioned in the caption. \textbf{O1/O2:} The laptop and table provide additional visual and textual content. \textbf{O3:} The caption omits the visible chair, showing a difference in image--caption content coverage.}}}

\newpage
\begin{validationprompt}{ObjectBench validation example: \mbox{\textit{[orange]}}}
\label{prompt:validation_orange}
\validationsystemprompt

\medskip
\hrule height 0.3pt
\medskip
\textbf{User prompt}\\
Annotate the following image--text pair.

\textbf{Input}\par
\noindent\begin{minipage}[t]{0.38\linewidth}
\vspace{0pt}
\includegraphics[width=\linewidth,height=0.18\textheight,keepaspectratio]{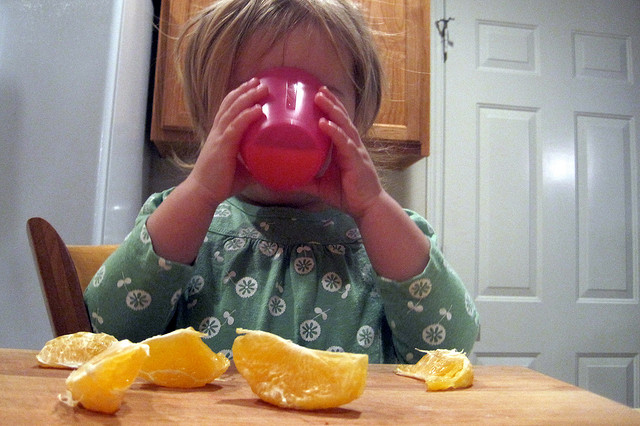}
\end{minipage}\hfill
\begin{minipage}[t]{0.58\linewidth}
\vspace{0pt}
\setlength{\parskip}{4pt}
Target concept: \textit{[orange]}\\
Concept definition: The COCO category orange: the citrus fruit, whole or in pieces.

Image: The original image shown on the left.

Caption: ``A little girl is drinking milk from a cup.''
\end{minipage}\par
\smallskip

\validationquestions{orange}

\textbf{Response}\\
Record \texttt{YES}, \texttt{NO}, or \texttt{UNCERTAIN} for O1--O3, and mark U1 and U2 as N/A. Use \texttt{UNCERTAIN} when the relevant content cannot be determined. Evaluators annotate independently without seeing one another's labels. Their responses are then combined in the table below.
\validationresponses{
\rowcolor{ValidationHumanA} Human annotator A & \texttt{N/A} & \texttt{N/A} & \texttt{YES} & \texttt{YES} & \texttt{UNCERTAIN} \\
\rowcolor{ValidationHumanB} Human annotator B & \texttt{N/A} & \texttt{N/A} & \texttt{YES} & \texttt{YES} & \texttt{YES} \\
\rowcolor{ValidationModelA} GPT-5.6 Sol & \texttt{N/A} & \texttt{N/A} & \texttt{YES} & \texttt{YES} & \texttt{YES} \\
\rowcolor{ValidationModelB} Kimi-K3 & \texttt{N/A} & \texttt{N/A} & \texttt{YES} & \texttt{YES} & \texttt{YES} \\
}
\end{validationprompt}

{\captionsetup{font=footnotesize,justification=justified,singlelinecheck=false,skip=0pt}
\captionof*{figure}{{\textbf{Prompt~\ref{prompt:validation_orange}.} An illustrative image--text pair for \textit{[orange]}. \textbf{U1/U2:} These questions are not applicable to \textit{ObjectBench}. \textbf{O1/O2:} The child, cup, and table provide additional visual content, while the child, cup, and milk provide additional textual content. \textbf{O3:} The caption omits the visible orange segments and table and mentions milk that is not visible inside the cup, showing differences in image--caption content coverage.}}}

\clearpage
\newpage
\bibliographystyle{plain}
\bibliography{paper}

\end{document}